\documentclass{article} %
\usepackage{iclr2027_conference_preprint,times}

\usepackage{amsmath,amsfonts,bm}

\newcommand{\xmark}{\ding{55}}

\def\eqref#1{equation~\ref{#1}}

\def\1{\bm{1}}

\DeclareMathAlphabet{\mathsfit}{\encodingdefault}{\sfdefault}{m}{sl}
\SetMathAlphabet{\mathsfit}{bold}{\encodingdefault}{\sfdefault}{bx}{n}

\usepackage[utf8]{inputenc} %
\usepackage[T1]{fontenc}    %
\usepackage{url}            %
\usepackage{booktabs}       %
\usepackage{amsfonts}       %
\usepackage{nicefrac}       %
\usepackage{microtype}      %
\usepackage{xcolor}         %
\usepackage{times}
\usepackage[inline]{enumitem}
\usepackage{graphicx}
\usepackage{tabularx}
\usepackage{amssymb}
\usepackage{pifont}
\usepackage{subcaption}
\usepackage{tikz}
\usetikzlibrary{arrows.meta, positioning, shapes.geometric, calc}
\usepackage{comicneue} %
\usepackage{upgreek}
\usepackage[most]{tcolorbox} %
\usepackage{multirow}

\usepackage{algorithm}
\usepackage[noend]{algpseudocode} %
\usepackage[pagebackref=true,breaklinks=true,colorlinks,bookmarks=false,citecolor=blue,linkcolor=blue]{hyperref}      %
\usepackage{cleveref} %

\newcommand{\rset}{\mathbb{R}}

\newcommand{\rmd}{\mathrm{d}}
\newcommand{\Id}{\mathrm{Id}}

\def\ie{\textit{i.e.,}}

\title{Adversarial Learning of Classifier-Free Guidance Schedules}

\author{%
  Ashwini Pokle\thanks{Equal contribution.} \\
  Google \\
  \texttt{apokle@google.com} \\
  \And
  Alexandre Galashov\footnotemark[1] \\
  Google DeepMind, Gatsby UCL \\
  \texttt{agalashov@google.com} \\
  \And
  Arnaud Doucet \\
  Google DeepMind \\
  \texttt{arnauddoucet@google.com} \\
  \And
  Mauricio Delbracio \\
  Google \\
  \texttt{mdelbra@google.com} \\
  \And
  Valentin De Bortoli \\
  Google DeepMind \\
  \texttt{vdebortoli@google.com} \\
}

\iclrfinalcopy %
\begin{document}

\maketitle

\begin{abstract}
Modern text-to-image diffusion models rely on classifier-free guidance (CFG) to achieve high image fidelity and text alignment.
However, CFG typically applies a static, global scale across all timesteps, samples, and conditions — a choice that is generally suboptimal and can introduce artifacts, as different states may benefit from different levels of guidance.
While time-varying schedules are known to improve quality, designing them by hand is non-trivial and application-dependent.
In this paper, we learn the guidance schedule as a function of diffusion time, conditioning and the current noisy sample, in order to better align sampled images with the text prompt. We frame this as a density ratio estimation problem: a discriminator is trained to estimate the time-dependent log-density ratio between the true and guided marginal distributions, while a lightweight generator network predicts the optimal, state-dependent guidance scale.
Empirically, our approach outperforms both heuristic CFG schedules and prior methods for learning dynamic guidance on text-to-image generation benchmarks.
\end{abstract}

\section{Introduction}
Diffusion models~\citep{song2019generative,song2020denoising,song2020score} and flow matching models~\citep{lipman2022flow} have emerged as state-of-the-art generative models for high-fidelity image and video generation. The impressive practical performance of these models is largely driven by the use of Classifier-Free Guidance (CFG)~\citep{ho2022classifier} during the sampling process. 

For flow matching, CFG replaces the conditional velocity approximation $v_{\theta}(x_t, t, c)$ with a linear combination of conditional and unconditional approximations; \ie~$(1+\omega) v_{\theta}(x_t, t, c) - \omega v_{\theta}(x_t, t, \varnothing)$, where $\omega$ is the guidance weight. 
A value of $\omega=0$ corresponds to conditional sampling, while $\omega=-1$ corresponds to the unconditional one. In practice, most of modern text-to-image (T2I) applications employ a constant guidance weight $\omega\approx 7.5$~\citep{rombach2022highresolutionimagesynthesislatent}.  Using such high guidance weight increases prompt alignment (i.e. CLIP score) and improves image quality.

Relying on a constant guidance weight $\omega$ can lead to artifacts and over-saturation in the generated images~\citep{saharia2022photorealistic,kynkaanniemi2024applying,sadat2024eliminating}.
Manually designed dynamical guidance schedules~\citep{kynkaanniemi2024applying,wang2024analysis} have been shown to mitigate these issues. However, these schedules introduce hyperparameters that can be difficult to tune across datasets, and are typically agnostic to the specific conditioning, applying the same guidance scale regardless of the complexity of text prompt or the specific class label.

Recently, efforts have shifted towards learning optimal guidance schedules that adapt to different conditioning information, thereby offering greater flexibility. 
~\citet{galashov2025learn} demonstrate that it is possible to learn time- and conditioning-dependent guidance weights that improve the Fr\'echet Inception Distance (FID)~\citep{fid} relative to constant CFG and manually designed CFG schedule baselines while maintaining comparable text alignment. Their approach is motivated by \emph{marginal consistency condition}
which encourages marginals of the true noising and the guided denoising distributions to match. Their practical implementation, however, relied on a stronger (in a sense that it implies \emph{marginal consistency}), \emph{self-consistency} objective which is while stable to train, failed to surpass constant CFG in text alignment.

In this paper, we 
shift our focus from satisfying the strict self-consistency condition to the weaker \textit{marginal consistency} condition. 
Unlike prior work that operates on raw latents via an energy distance kernel~\citep{SzeRiz04} with maximum mean discrepancy (MMD)~\citep{gretton2012kernel}, we employ an adversarial framework to learn the guidance schedule by matching the marginal distribution of the guided trajectories to that of the true data distribution at every timestep.

We demonstrate that this condition can be formulated as an objective that involves density ratio estimation, which naturally fits into a generative adversarial network (GAN) framework~\citep{goodfellow2014generative}.
In our setup, a discriminator estimates the log-density ratio between real and classifier-free guided samples, while a generator learns to predict guidance scales that maximize this ratio.
By optimizing for marginal consistency across the sampling trajectory, our method achieves good prompt adherence as well as sample quality, outperforming constant guidance and learned guidance baselines in T2I benchmarks.

Our main contributions are summarized as follows. First, we propose a method to learn guidance schedules based on the marginal consistency condition, encouraging the distribution of guided samples to match the true data distribution at every timestep along the sampling trajectory. Second, we use an adversarial framework to estimate density ratios and learn dynamic time-, conditioning-, and noisy sample-dependent guidance scales. Third, we show that our method outperforms constant and manually selected guidance schedule baselines as well as prior learned schedules on T2I benchmarks.

\section{Background}

\textbf{Notation.} 
Following the flow matching (FM) convention~\citep{lipman2022flow}, our framework uses continuous time $t \in [0, 1]$, where $t=0$ is pure noise and $t=1$ is data. Denoising steps proceed from a proposal timestep $s$ to a target timestep $t$ ($s<t$). Here, $p_t$ denotes the true marginal distribution of the noised data. We also use $p_{t|s}$ to denote the conditional distribution $p_{t|s}(x_t |x_s)$ for any $s$ and $t$; $p_{1|c}$ to denote $p(x_1|c)$; and $p_{1,c}$ to denote the joint distribution $p(x_1,c)$. We write $p_t^{s, \boldsymbol{\omega}}(x_t)$ for the marginal distribution of the guided particles, obtained by taking a sampling step from time $s$ to $t$ using the dynamically guided vector field (see definitions below). We use the bold notation $\boldsymbol{\omega}$ to denote the schedule of the guidance weights. Finally, $\mathcal{N}(\mu,\Sigma)$ denotes the Gaussian distribution with mean $\mu$ and covariance $\Sigma$, and $\mathcal{N}(x;\mu,\Sigma)$ represents its density evaluated at $x$.

\textbf{Flow matching.}
  The goal of conditional generative modeling is to sample from a target conditional distribution $p_{1|c}$ on $\rset^d$, where $c \sim p(c)$ is a conditioning signal (e.g., a text prompt). To achieve this, Flow matching~\citep{lipman2022flow} constructs a probability path connecting a standard Gaussian $p_0 = \mathcal{N}(0, \Id)$ at $t=0$ to the data distribution $p_1 = p_{\mathrm{data}}$ at $t=1$ by using
  \begin{equation}
  \label{eq:interpolation}
    \textstyle
      x_t = (1-t)\, \varepsilon + t\, x_1, \quad \varepsilon \sim \mathcal{N}(0, \Id), \quad x_1 \sim p_{1|c}.
  \end{equation}
  This induces the conditional distribution $p_{t|1}(x_t | x_1) = \mathcal{N}(x_t;\, t\,x_1,\, (1-t)^2 \Id)$.
  The conditional velocity field generating this path is $u_t(x_t | x_1) = x_1 - \varepsilon$.
  A neural network $v_\theta(x_t, t, c)$ is trained to approximate the marginal velocity field by minimizing the following objective
  \begin{equation}
  \textstyle
  \label{eq:flow_matching_loss}
      \mathcal{L}(\theta) = \mathbb{E}_{t \sim \mathcal{U}[0,1],\, (x_1, c) \sim p_{1,c},\, \varepsilon \sim \mathcal{N}(0,\Id)} \left[ \left\| v_\theta(x_t, t, c) - (x_1 - \varepsilon) \right\|^2 \right].
  \end{equation}
  
  \textbf{Conditional sampling.}
  At inference time,  the samples are generated by integrating the following the ordinary differential equation (ODE) with learned velocity field starting from $x_0 \sim \mathcal{N}(0, \Id)$,  
  \begin{equation}
  \label{eq:ode}
      \rmd x_t= v_\theta(x_t, t, c) \rmd t , 
  \end{equation}
  from $t=0$ to $t=1$. Using Euler solver and time discretization $0 = t_0 < t_1 < \dots < t_N = 1$, we get
  \begin{equation}
  \label{eq:euler_step}
        \textstyle
      x_{t_{k+1}} = x_{t_k} + (t_{k+1} - t_k) \, v_\theta(x_{t_k}, t_k, c), \quad k = 0, \dots, N-1.
  \end{equation}
  For a single step from time $s$ to time $t > s$, the update is
  $x_t = x_s + (t - s)\, v_\theta(x_s, s, c)$.
  The predicted clean data from a noisy sample $x_t$ can be recovered as
  \begin{equation}
  \label{eq:denoiser}
        \textstyle
      \hat{x}_1(x_t, t, c) = x_t + (1-t) \, v_\theta(x_t, t, c).
  \end{equation}
  
  \textbf{Classifier-Free Guidance (CFG).} Let $v_\theta(x_t, t, \varnothing)$ be the unconditional velocity (trained by randomly dropping the conditioning) and $\omega$ is the guidance weight.
  Classifier-Free Guidance~\citep{ho2022classifier} replaces the conditional velocity with the guided velocity
  \begin{equation}
  \label{eq:cfg_velocity}
      \textstyle
      v_\theta^{\mathrm{cfg}}(x_t, t, c; \omega) = v_\theta(x_t, t, c) + \omega \left( v_\theta(x_t, t, c) - v_\theta(x_t, t, \varnothing) \right),
  \end{equation}
  $\omega = -1$ yields unconditional sampling, $\omega = 0$ conditional sampling, while $\omega > 0$ amplifies the conditioning. Typically,   
  $\omega \approx 7.5$ is standard for 
  T2I applications~\citep{saharia2022photorealistic,rombach2022highresolutionimagesynthesislatent}. However, constant $\omega$ across time and conditioning may lead to artifacts
  such as over-saturation~\citep{kynkaanniemi2024applying,sadat2024eliminating}.
  
 \textbf{Dynamic guidance schedules.}
  A natural extension of CFG is to allow the guidance weight $\omega$ to vary across the sampling trajectory, i.e., $\omega = \omega(t)$.
  Several heuristic schedules have been proposed, including limiting guidance to a time interval (LIG;~\citealt{kynkaanniemi2024applying}), clamp-linear schedules (CLG;~\citealt{wang2024analysis}), and time-dependent rescaling~\citep{sadat2024no}.
  While these can improve over constant guidance, they still apply the same schedule globally across all samples and conditionings, and require manual tuning for each model and application.

  \textbf{Learnable guidance schedules and marginal consistency.} 
  Rather than hand-designing a schedule, 
  one can \emph{learn} guidance weights $\omega_{c,(s,t)}$ as functions of the conditioning $c$ and the timesteps $s,t$ with $s < t$ (time $t$ is the target and time $s$ is the source time).
  One could attempt to learn $\omega$ with guided velocity $v_\theta^{\mathrm{cfg}}(x_s, s, c; \omega)$~\eqref{eq:cfg_velocity} from~\eqref{eq:flow_matching_loss}. However, this will necessarily lead to $\omega = 0$ since $v_{\theta}$ is already a minimizer of~\eqref{eq:flow_matching_loss}.
  Recent work in learned guidance has explored enforcing \emph{consistency conditions} to learn $\omega$. 
  Let the true marginal at time $t$ be given by
  \begin{equation}
  \label{eq:true_marginal}
      p_t(x_t) = \int p_{t|1}(x_t | x_1)\, p_{1,c}(x_1, c)\, \rmd x_1\, \rmd c . 
  \end{equation}
  Sampling $(x_1, c) \sim p_{1,c}$, noising to time $s$ via $x_s \sim p_{s|1}(\cdot | x_1)$, and taking a guided step from $s$ to $t$ using~(\eqref{eq:cfg_velocity}) with weight $\omega$, will lead to a \emph{guided marginal}
  \begin{equation}
  \label{eq:guided_marginal}
      p_t^{s,\boldsymbol{\omega}}(x_t) = \int \left[ \int p_{t|s,c}^{(\theta,\boldsymbol{\omega})}(x_t | x_s, c)\, p_{s|1}(x_s | x_1)\, \rmd x_s\, \right]p_{1,c}(x_1, c)\, \rmd x_1\, \rmd c,
  \end{equation}
  where $p_{t|s,c}^{(\theta,\boldsymbol{\omega})}(x_t | x_s, c)$ is the guided transition kernel from time $s$ to $t$; for a deterministic Euler solver, $p_{t|s}^{(\theta,\boldsymbol{\omega})}(x_t | x_s, c) = \updelta\!\left(x_t - x_s- (t - s)\, v_\theta^{\mathrm{cfg}}(x_s, s, c;\, \boldsymbol{\omega})\right)$.
  We say that the guided process satisfies \emph{marginal consistency} if, for all $0 \leq s < t \leq 1$,
  \begin{equation}
  \label{eq:marginal_consistency}
      p_t^{s,\boldsymbol{\omega}}(x_t) \approx p_t(x_t).
  \end{equation}
  \citet{galashov2025learn} enforced a stronger \emph{self-consistency} variant of~\eqref{eq:marginal_consistency}---matching $p_{t|1,c}$ and $p^{s,\boldsymbol{\omega}}_{t|1,c}$ (the inner integral in~\eqref{eq:guided_marginal})---using MMD~\citep{gretton2012kernel} with energy kernel~\citep{SzeRiz04}, achieving strong FID improvements but not consistently outperforming constant guidance on text alignment metrics. They argued that the gradients of the objective based on \emph{marginal consistency} suffer from high variance due to a marginalization over $(x_1,c)$ and the use of the MMD with a fixed energy kernel defined on raw high dimensional data. In the next section, we propose an adversarial approach to enforce marginal consistency via density ratio estimation, yielding superior text-to-image performance.

\section{Learning to Guide with Density Ratio Estimation}

\begin{figure}[h]
    \centering
\includegraphics[width=.7\textwidth]{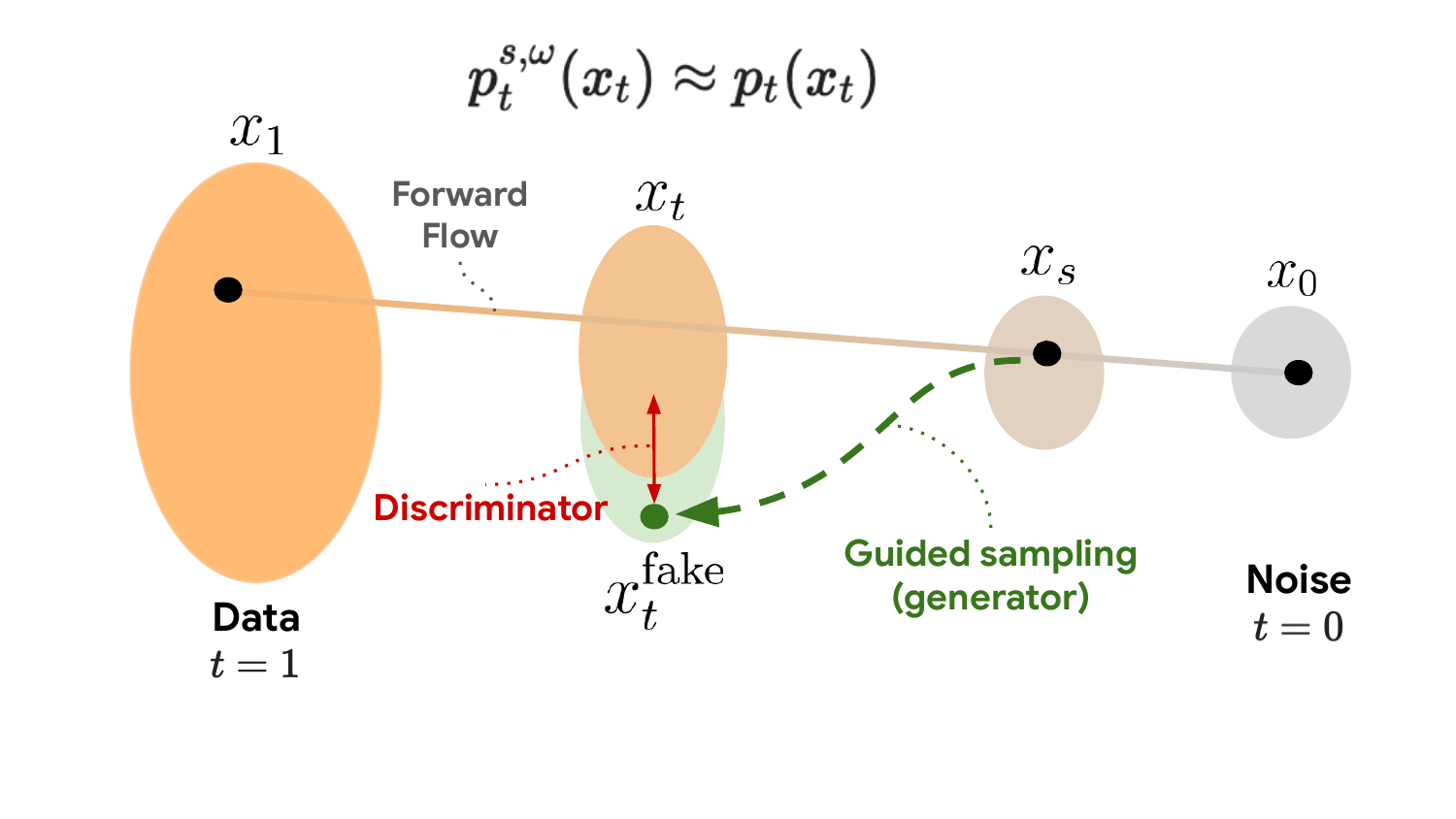}
    \caption{Marginal consistency via GAN. We draw two independent samples, $x_1^1$ and $x_1^2$, from the data distribution. First sample $x_1^1$ is noised to the time $s$ and then denoised with guidance to the time $t$ which denotes our generated path. Second sample $x_1^2$ is noised to the time $t$ through forward process and denotes the real samples. The discriminator is then trained to discriminate between these two.
    }
    \label{fig:diagram}
\end{figure}

We introduce our method for learning the dynamic guidance weights $\boldsymbol{\omega}$ for diffusion sampling with a generative adversarial network (GAN). 
To enforce the \emph{marginal consistency} condition, we derive an adversarial objective based on density ratio estimation using a learned discriminator. 
We assume access to a pre-trained conditional flow matching model, $v_{\theta}(x_t, t, c)$, and an unconditional model, $v_{\theta}(x_t, t, \varnothing)$.

\subsection{Guidance learning objective}

In standard CFG~(\eqref{eq:cfg_velocity}), a static scalar $\omega$ controls the guidance strength uniformly across all timesteps and inputs. %
We parametrize the guidance weights to be a function of the times $(s,t)$, the conditioning $c$, and the current noisy sample $x_s$
i.e. ${\boldsymbol{\omega}}_{x_s,c,(s,t)}$. 
Intuitively, adding more information to the guidance weights might allow the model to more finely adjust the guidance weights and adapt to a specific noise trajectory.

\textbf{Adversarial objective.} For $(s,t) \sim p(s, t)$ (see Section~\ref{sec:guidance_learning_algo} for more details), the \emph{marginal consistency} condition dictates $p_t^{s, \boldsymbol{\omega}}(x_t) \approx p_t(x_t)$, i.e. the distribution of our guided particles should match the true data distribution.
This can be achieved by minimizing the Kullback-Leibler (KL) divergence
\begin{equation}
    \textstyle
    \mathbb{E}_{(s, t) \sim p(s,t)}\left[\mathrm{KL}\left(p_t^{s, \boldsymbol{\omega}}(x_t)  \| p_t(x_t)\right)\right] .
\end{equation}
This divergence can be rewritten in terms of the density ratio $r^{\boldsymbol{\omega}}_t(x_t) =p_t(x_t)/p_t^{s, \boldsymbol{\omega}}(x_t)$ as:
\begin{equation}
    \textstyle
    \mathrm{KL}\left(p_t^{s, \boldsymbol{\omega}}(x_t)  \| p_t(x_t)\right)  = -\mathbb{E}_{p_t^{s, \boldsymbol{\omega}}}\left[\log r_t^{\boldsymbol{\omega}}(x_t)\right].
\end{equation}
We train a generator to predict $\boldsymbol{\omega}$ to minimize this divergence and a discriminator to estimate this log-density ratio, providing training signal to the generator. The approach is illustrated in Figure~\ref{fig:diagram}. 

\textbf{Discriminator.} We define a discriminator $d_\phi(x_t, s, t, c)$ with parameters $\phi$ which needs to distinguish between true marginal samples $x_t^{\text{real}} \sim p_t(x_t)$ and guided marginal samples $x_t^{\text{fake}}(\boldsymbol{\omega}) \sim p_t^{s, \boldsymbol{\omega}}(x_t)$. We optimize this via a standard binary cross-entropy loss (minus sign due to maximization):
\begin{equation}
    \textstyle
    \mathcal{L}_{disc}(\phi) = -
    \left\{ \mathbb{E}_{p_t} \left[ \log \sigma(d_\phi) \right] + \mathbb{E}_{p_t^{s, \boldsymbol{\omega}}} \left[ \log (1 - \sigma(d_\phi)) \right] \right\},
\end{equation}
where $\sigma(\cdot)$ is the sigmoid function. At convergence, the optimal discriminator $d^\star$ is a Bayes classifier. Extracting the logit from this optimal discriminator directly recovers the log-density ratio:
\begin{equation}
    \textstyle
    \log \left(\dfrac{d^\star}{1 - d^\star}\right) = \log \left(\dfrac{p_t(x_t)}{p_t^{s, \boldsymbol{\omega}}(x_t)}\right) = \log r_t^{\boldsymbol{\omega}} (x_t)  \approx d_\phi(x_t, s, t, c) . 
\end{equation}

We apply a standard $R_1$ gradient penalty~\citep{mescheder2018training} to the discriminator to ensure smooth ratio estimation with a constant $\gamma > 0$ (hyperparameter): 
\begin{equation}
    \mathcal{L}_{\text{reg}}(\phi) = \frac{\gamma}{2} \mathbb{E}_{x_t^{\text{real}}} [\|\nabla_{x_t} d_\phi(x_t^{\text{real}}, s, t, c)\|^2].
\end{equation}

The total loss for the discriminator is
\begin{equation}
    \textstyle\mathcal{L}_D(\phi) = \mathcal{L}_{disc}(\phi) - \frac{\gamma}{2} \mathbb{E}_{x_t^{\text{real}}} [\|\nabla_{x_t} d_\phi(x_t^{\text{real}}, s, t, c)\|^2]
    \label{eq:disc_total_loss}
\end{equation}

\textbf{Generator.} We define a generator or a guidance network $\boldsymbol{\omega}^{\psi}_{x_s,c,(s,t)} = \boldsymbol{\omega}(x_s,s,t,c;\psi) > 0$ with parameters $\psi$, which outputs non-negative guidance weights. Employing these weights via guided velocity~\eqref{eq:cfg_velocity} specifies guided marginal distribution $p_t^{s, \boldsymbol{\omega}}(x_t)$~\eqref{eq:guided_marginal}. We write here $x_t^{\text{fake}}(\psi) \sim p_t^{s, \boldsymbol{\omega}}$ to highlight the dependence on parameters $\psi$. The generator is trained by optimizing
\begin{equation}
    \textstyle
    \mathcal{L}_{gen}(\psi) = 
    \mathbb{E}_{p_t^{s, \boldsymbol{\omega}}} \left[ -d_\phi(x_t^{\text{fake}}(\psi), s, t, c) \right].
\end{equation}

To prevent the guidance scale from exploding or over-correcting out of distribution, we introduce a stabilizing $L_2$ penalty to the generator loss with  time-dependent regularization weight $\lambda(t)$,
\begin{equation}
    \label{eq:reg_term}
    \textstyle
    \mathcal{R}({\boldsymbol{\omega},t}) = \lambda(t) \boldsymbol{\omega}^2 . 
\end{equation}
In practice, we set $\lambda(t) = at^2$, where $a$ is a hyperparameter.

\textbf{Reward-based regularization.}
Similar to prior work \citep{galashov2025learn}, we introduce a reward-based loss term. We compute the first-order approximation of the clean data using $v_t^{\text{cfg}, \boldsymbol{\omega}}$ as
\begin{equation}
    \textstyle
\hat{x}_1(\boldsymbol{\omega}) = x_s + (1 - s) \cdot v_t^{\text{cfg}, \boldsymbol{\omega}}(x_s,s,1,c;\psi)
\end{equation}
We define the reward loss as the negative expectation of a practitioner-defined reward function $R(\hat{x}_1, c)$, such as the CLIP score for text-to-image models,
\begin{equation}
\textstyle
\mathcal{L}_{\text{Reward}}(\boldsymbol{\omega}) = -\mathbb{E} \left[ R(\hat{x}_1(\boldsymbol{\omega}), c) \right].
\label{eq:clip_reward_loss}
\end{equation}
The total loss for the generator is given as
\begin{align}
\label{eq:gen_loss_tot}\textstyle\mathcal{L}_{\text{G}}(\psi) = \mathcal{L}_{gen}(\psi)  + \mathcal{R}({\boldsymbol{\omega},t}) + \eta \mathcal{L}_{\text{Reward}}(\boldsymbol{\omega}),
\end{align}
where $\eta>0$ is a hyperparameter. In our framework, adversarial loss and regularization loss terms ensure that the samples remain on the data manifold while the reward term pushes for better alignment.

\subsection{Guidance Learning Algorithm Settings}\label{sec:guidance_learning_algo}

  \textbf{Time distribution $p(s, t)$.} We use the same time distribution $p(s,t)$ as previously leveraged in \citet{galashov2025learn} in our experiments, and we re-write it in a flow matching notation. The target time is given by $t \sim \mathcal{U}[\zeta + \delta,\, 1 - \zeta]$, where $\zeta$ is a small constant which serves as boundary buffers that prevent optimization from encountering numerical singularities or vanishing signals found at extreme noise or pure data limits. In our experiment we set it to small constant $(10^{-2})$ without further tuning.  The step size between the source and target timesteps is $\Delta s \sim \mathcal{U}[\delta,\, t - \zeta]$. Therefore, the source time is $s = t - \Delta s$. Similar to prior observations, we observe that $\delta = 0.1$ slightly outperforms $\delta = 0.01$, even if inference utilizes smaller steps. See~\Cref{tab:mmdit_delta_study} for an ablation on $\delta$. This is likely because a larger $\delta$ provides a more robust signal to the discriminator.
To ensure that the generator interpolates smoothly to the smaller step sizes utilized at inference, we make our guidance weight predictor more robust by conditioning on coordinate-invariant geometric statistics rather than raw high dimensional latents and Fourier embeddings for timesteps. See our discussion on network parametrization below. In Appendix~\ref{app_sec:experimental_details}, we discuss alternative choices of time distribution.

\textbf{Network Parametrization.} 
The discriminator $d_\phi(x_t, s, t, c)$ is a convolutional network with parameters $\phi$.
The guidance network $\boldsymbol{\omega}(x_s,s,t,c;\psi)$
is a lightweight multi-layer perceptron (MLP) with parameters $\psi$. 
Because $x_s$ is high dimensional latent, feeding raw latents into the MLP is computationally inefficient. 
We instead compute a vector of summary statistics that captures the relationship between the current state, the conditioning, and the sampling trajectory.
Specifically, we extract the following statistics: 1) The log norm of the conditional velocity $\|v_\theta(x_s, s, c)\|$ and the CFG direction $\|v_\delta\|$, where $v_\delta = v_{\theta}(x_s, s, c) - v_{\theta}(x_s, s, \varnothing)$. 2) The cosine similarity between the current latent $x_s$ and the guidance direction $v_\delta$, calculated as $\frac{x_s \cdot v_\delta}{\|x_s\| \|v_\delta\|}$. This captures how well the current sample aligns with the intended conditioning signal. 3) The log distance to the data manifold $\log(1-s)$ and the log step size of the current jump $\log(t-s)$.
The MLP takes a concatenated feature vector with these values as an input. 
In addition, we find it useful to feed in Fourier embeddings of $s$, $t$ and $t-s$ to the generator. This helps the network interpolate smoothly to smaller step sizes utilized at inference.
We apply softplus activation at the output to ensure strictly non-negative guidance weights, mirroring the standard CFG behavior. 
For more architectural details, see~\Cref{app_sec:experimental_details}.

\textbf{Training strategy.}
The training of our guidance scheduler follows an adversarial paradigm that alternates between updating the discriminator and the generator \ie the guidance MLP. 
To ensure stable convergence and accurate density ratio estimation, we adopt the Two-Time-Scale Update Rule (TTUR) \citep{fid}, performing a single gradient step for each network alternately.

A key distinction in our training loop is the use of independent image-conditioning pairs, $(x_1^1, c^1)$ and $(x_1^2, c^2)$, drawn from the dataset $\mathcal{D}$. 
We use the first pair to generate the ``real'' marginal target $x_t^{\text{real}}$ and the second pair to define
the ``fake'' guided sample $x_t^{\text{fake}}$. 
This independent sampling is fundamental to enforcing marginal consistency; it prevents the discriminator from trivially learning to identify the specific noise or structure of a single trajectory. 
Instead, it forces the discriminator to evaluate whether the distribution of the guided particles matches the global data distribution at that specific timestep.
While this independent sampling is theoretically more aligned with the marginal consistency objective, we observe in practice that the framework is robust to this choice, yielding comparable performance even when correlated pairs are used.
See \Cref{tab:mmdit_xs_correlated_vs_independent} for quantitative results.

Putting together the above components, we summarize our method in \Cref{alg:flowmatching-gan-dre}.

\begin{algorithm}
    \begin{algorithmic}
        \Require Pretrained FM model $v_\theta$, a T2I dataset, time distribution $p(s,t)$, $\gamma$ discriminator regularization parameter, $\lambda$ generator regularization parameter, $\eta$ generator reward score parameter. Discriminator $d_\phi(x_t, s, t, c)$ with parameters $\phi$ and guidance network $\boldsymbol{\omega}(x_s,s,t,c;\psi)$ with parameters $\psi$. Learning rates $\alpha_{D}$ and $\alpha_{G}$ for discriminator and generator.
        \While{not converged}
        \State Sample times $s,t \sim p(s, t)$
        \State (\textbf{Real path}) Sample $(x_1^1, c^1) \sim \mathcal{D}$, $\epsilon \sim \mathcal{N}(0, I)$, $x_{t}^{\text{real}} = (1 - t)\epsilon + t x_1^1$
        \State Sample $(x_1^2, c^2) \sim \mathcal{D}$, $z \sim \mathcal{N}(0, I)$, $x_s = (1-s)z + s x_1^2$
        \State Compute $v^c_s = v_\theta(x_s, s, c^2)$ and  $v^u_s = v_\theta(x_s, s,\varnothing)$
        \State Compute $v_s^{\text{cfg}, \boldsymbol{\omega}}(\psi) = v_s^c + \boldsymbol{\omega}(x_s,s,t,c;\psi)(v_s^c - v_s^u)$ 
        \State (\textbf{Fake path}) $x_{t}^{\text{fake}}(\psi) = x_s + v_s^{\text{cfg}, \boldsymbol{\omega}}(\psi) (t - s)$ \Comment Euler/DDIM ODE sampling
        
        \State Update GAN discriminator $d_\phi$ by optimizing $\mathcal{L}_{D}(\phi)$~(\eqref{eq:disc_total_loss})
        \begin{align*}
            \phi \leftarrow\phi - \alpha_{D}\nabla  \mathcal{L}_{D}(\phi)
        \end{align*}

        \State Use the updated discriminator to update the generator $\psi$ by minimizing $\mathcal{L}_{G}(\psi)$~(\eqref{eq:gen_loss_tot})
        \State
        \begin{equation}
            \psi \leftarrow\psi -\alpha_{G}\nabla\mathcal{L}_{G}(\psi)
        \end{equation}
        \EndWhile
    \end{algorithmic}
    \caption{Algorithm to learn parameter of CFG with Flow Matching model.}
    \label{alg:flowmatching-gan-dre}
\end{algorithm}

\section{Related Work}

\textbf{Classifier-free guidance and Variants.} Classifier-Free Guidance (CFG)~\citep{ho2022classifier} has become the standard for conditional sampling in diffusion and flow models.
A significant amount of effort has been directed towards understanding the underlying mechanics of this success. \citet{chidambaram2024does} find that CFG might fail to sample from the intended tilted distribution, instead driving the trajectory towards "archetypes" of the class in the distribution. 
Further analysis in high-dimensional settings ~\citep{pavasovic2025understanding} and general data distributions \citep{li2025provable} has clarified how CFG alters the geometry of the score-field.
From a practical perspective, several training-free strategies have been proposed to mitigate oversaturation and artifacts associated with high constant guidance scales.
CFG++ \citep{chung2024cfg++} introduces manifold constraints to the sampling process, while \citet{sadat2024eliminating} decompose network predictions into parallel and orthogonal components. Autoguidance~\citep{karras2024guiding} rethinks the guidance direction by using a lower-quality version of the model as the unconditional reference.

Heuristic-based dynamic schedules have also seen rapid development. \citet{kynkaanniemi2024applying} apply guidance on a limited interval to improve image fidelity and alignment, while \citet{wang2024analysis} provide an extensive analysis of guidance dynamics. We adapt clamp-linear schedule from this work as a baseline.  \citet{malarz2025classifier} use the density of a beta distribution to modulate the guidance signal. 
Other training-free methods include rectified guidance ~\citep{xia2025rectified}, geo-guide~\cite{poleski2025geoguide}, and adaptive guidance~\citep{castillo2025adaptive}.
While these methods avoid the cost of training, they generally rely on global schedules that are agnostic to the specific semantic complexity of the conditioning.

Our work belongs to the emerging category of learned guidance schedules. Unlike \citet{felix2025feedback} which uses internal model predictions; \ie feedback during inference, we optimize a dedicated guidance scheduler. 
Closely related to our work is MMD guidance~\citep{galashov2025learn}. The key distinction is that we enforce a weaker marginal consistency condition instead of the strict self-consistency condition in an adversarial framework.

Another line of work that is related to our work is discriminator guidance~\citep{kim2022refining, verine2025improving}. 
These methods use a discriminator as corrector to provide auxiliary score gradient, thereby increasing inference-time complexity. On the other hand, we use the discriminator strictly as a training supervisor to learn optimal guidance schedule. 
As our generator is a lightweight MLP, our method can achieve better alignment with the prompts while maintaining the sampling efficiency of the original backbone.

\textbf{Adversarial training in Generative Modeling.} 
Our method leverages the principles of Generative Adversarial Networks~\citep{goodfellow2014generative} to learn dynamic guidance schedules.
GANs were traditionally used as standard alone generative models, but later on they have also been integrated into likelihood-based settings, especially in the context of diffusion models, to enhance sampling efficiency and quality. 
Diffusion GAN~\citep{wang2022diffusion} and Denoising Diffusion GAN~\citep{xiao2021tackling} use a discriminator to match the conditional or marginal distributions of denoising steps, thereby speeding up sampling while maintaining  visual fidelity.
In this work, we do not use the adversarial loss to train the primary generative backbone. Instead, we use it to provide a training signal for the guidance scheduler.

\section{Experiments}
\subsection{Experimental Setup}\label{sec:experimental_setup}

\textbf{Datasets and Models.} We evaluate the performance of our method on the standard text-to-image generation task using MS-COCO 2014 dataset~\citep{lin2014microsoft} at $512\times512$ resolution. 
Our experiments utilize two sizes of a custom-trained flow matching model~\citep{lipman2022flow} based on the Multimodal Diffusion Transformer backbone~\citep{labs2025flux1kontextflowmatching} with parameter counts of 740M (MMDiT-XS) and 1.05B (MMDiT-S). During all subsequent training of the guidance network, this remains frozen. For CFG, instead of using empty conditioning $\varnothing$ in~\eqref{eq:cfg_velocity}, we use a fixed negative prompt $c_{\text{neg}} =$ ``blurred, blurry, disfigured, ugly, tiling, poorly drawn'' and therefore replace the unconditional velocity term $v_\theta(x_t, t, \varnothing)$ by $v_\theta(x_t, t, c_{\text{neg}})$. We found that it led to higher quality images than $\varnothing$ and we use it for our learned models and the baselines. We provide detailed hyperparameter settings for optimizer as well as details of the model architecture in \Cref{app_sec:experimental_details}.

\textbf{Baselines.}
We evaluate our method against a diverse set of guidance strategies:
1) \textit{Fixed Guidance}: We consider unguided ($\omega=0$) and the standard constant CFG ($\omega=7.5$) as our primary foundational baselines. 
2) \textit{Heuristic-based guidance schedules}: We evaluate state-of-the-art manually designed schedules, including Limited Interval Guidance (LIG) \citep{kynkaanniemi2024applying} and Clamp-linear schedules (CLG) \citep{wang2024analysis}. For these baselines, we perform extensive hyperparameter tuning to find the optimal ranges for MS-COCO, as detailed in \Cref{sec:baseline_ablations}. 
3) \textit{Learned Self-Consistency (MMD+SC) / Marginal-Consistency (MMD+MC)}: We compare against the approach by \citet{galashov2025learn}, which utilizes Maximum Mean Discrepancy and an energy kernel to enforce self-consistency / marginal-consistency.

\textbf{Evaluation Metrics.}
We employ a suite of metrics to better capture the trade-off between image fidelity and text alignment. We use FID \citep{fid} to measure the distributional distance between generated and real images. We report the CLIP score \citep{clip_score,radford2021learning} to measure alignment between the generated images match the input prompts. We include LAION Aesthetics metric~\citep{schuhmann2022laionaesthetics} to capture visual quality. Finally, we also report Human Preference Score v2 (HPSv2)~\citep{wu2023human} and PickScore~\citep{kirstain2023pick} to measure alignment with human preferences, and prompt adherence.

\subsection{Text-to-Image (T2I) Generation}
\label{sec:t2i_generation}

\begin{table}[htbp!]
\centering
\caption{\textbf{MS COCO $512 \times 512$}. Performance of different methods on a small, MMDiT-XS model. We highlight the best metric in \textbf{bold}.
}
\resizebox{\textwidth}{!}{
\begin{tabular}{ccccccc}
\toprule
\textbf{Method} & \textbf{Guidance Weight} & \textbf{FID $\downarrow$} & \textbf{CLIP $\uparrow$} & \textbf{Aesthetic $\uparrow$} & \textbf{HPSv2 $\uparrow$} & \textbf{PickScore $\uparrow$} \\
\midrule
& \textbf{Baselines} \\
\midrule
Unguided & $\omega = 0$ & 27.65 & 0.2742 & 4.79 & 0.2547 & 0.1997 \\
Constant & $\omega = 7.5$ & 29.73 & 0.3044 & 5.29 & 0.2817 & 0.2186 \\
LIG& $\omega(t) = 12.5,  t \in [0.1, 0.9]$ & \textbf{25.57} & 0.3032 & 5.30 & 0.2812 & 0.2171 \\ %
CLG & $\omega = 15.0$ & 26.38 & 0.3038 & 5.26 & 0.2807 & 0.2175 \\
MMD + SC & $\boldsymbol{\omega}(x_s,s,t,c)$ & 28.38 & 0.30419 & 5.2647 & 0.28129 & 0.21813 \\
MMD + MC & $\boldsymbol{\omega}(x_s,s,t,c)$ & 28.10983 & 0.30368 & 5.25155 & 0.28074 & 0.21789 \\
\midrule
GAN + MC & $\boldsymbol{\omega}(x_s,s,t,c)$ & 31.31 & \textbf{0.3048} & \textbf{5.32} & \textbf{0.2829} & \textbf{0.2186} \\
\bottomrule
\end{tabular}
}
\label{table:ms-coco-results-small}
\end{table}

\begin{table}[htbp!]
\centering
\caption{\textbf{MS COCO $512 \times 512$}. Performance of different methods on a large, MMDiT-S model. We highlight the best metric in \textbf{bold}.
}
\resizebox{\textwidth}{!}{
\begin{tabular}{ccccccc}
\toprule
\textbf{Method} & \textbf{Guidance Weight} & \textbf{FID $\downarrow$} & \textbf{CLIP $\uparrow$} & \textbf{Aesthetic $\uparrow$} & \textbf{HPSv2 $\uparrow$} & \textbf{PickScore $\uparrow$} \\
\midrule
& \textbf{Baselines} \\
\midrule
Unguided & $\omega = 0$ & 24.3 & 0.279 & 4.84 & 0.2588 & 0.2024 \\
Constant & $\omega = 7.5$ & 31.11 & 0.3063 & 5.31 & 0.285 & \textbf{0.2214} \\
LIG & $\omega(t) = 15.0, t \in [0.1, 0.9]$ & \textbf{24.95} & 0.3042 & 5.31 & 0.2843 & 0.2194 \\
CLG& $\omega = 14.0$ & 27.14 & 0.3055 & 5.27 & 0.2838 & 0.2203 \\
MMD + SC & $\boldsymbol{\omega}(x_s,s,t,c)$ & 28.54873 & 0.30674 & 5.29430 & 0.28471 & 0.22078 \\
MMD + MC & $\boldsymbol{\omega}(x_s,s,t,c)$ & 29.43209 & 0.30625 & 5.27623 & 0.28364 & 0.22049 \\
\midrule
GAN + MC & $\boldsymbol{\omega}(x_s,s,t,c)$ & 31.7 & \textbf{0.3068} & \textbf{5.34} & \textbf{0.28564} & \textbf{0.2214} \\
\bottomrule
\end{tabular}
}
\label{table:ms-coco-results-large}
\end{table}

\textbf{Results.} 
We summarize the quantitative performance of our method on MS-COCO 512$\times$512 benchmark in \Cref{table:ms-coco-results-small} for MMDiT-XS and in \Cref{table:ms-coco-results-large} for MMDiT-S. We used the same generator architecture for MMD+SC/MMD+MC baseline and we add more extensive comparisons in~\Cref{sec:extensive_comparisons}, where we replicated the same setup as in~\citep{galashov2025learn}. Across both model sizes, our approach based on adversarial framework that enforces marginal consistency (GAN + MC) consistently achieves better metrics compared to the baselines in almost all alignment and human preference metrics.
Specifically, our method yields superior CLIP, Aesthetic, and HPSv2 scores compared to constant guidance and manually tuned heuristic schedules. 

As shown in \Cref{table:ms-coco-results-small}, MMDiT-XS model benefits significantly from our learned schedule. While the LIG achieves the lowest FID (25.64), it does so at the cost of lower alignment. Our method achieves a CLIP score of 0.305 and an Aesthetic score of 5.32, the highest in the group. 
The gains observed in the smaller model translate effectively to the larger MMDiT-S architecture as shown in \Cref{table:ms-coco-results-large}. GAN + MC outperforms baselines on CLIP (0.3068), Aesthetic (5.34), and HPSv2 (0.2856).
This indicates that our approach scales with model capacity, providing a more reliable guidance signal than manually engineered schedules.

This suggests that learning a dynamic, conditioning-dependent schedule via marginal consistency allows the model to better navigate the trade-off between image realism and adherence to complex text prompts.
Our approach also demonstrates a clear advantage over self-consistency framework of \citet{galashov2025learn} on alignment and human preference metrics.
Finally, as observed in \Cref{table:ms-coco-results-small,table:ms-coco-results-large}, our method also results in a marginal increase in FID. 
This increase is a known characteristic of adversarial training and CLIP-reward optimization, where the model prioritizes semantic features and visual pop over the matching of low-level Inception-v3 statistics.
This increase also highlights the known misalignment between Inception-v3 statistics and human aesthetic preferences in high guidance regimes ~\citep{jayasumana2024rethinking}.

\begin{figure}[h]
    \centering
    \begin{minipage}[c]{0.4\textwidth}
    \includegraphics[width=\textwidth]{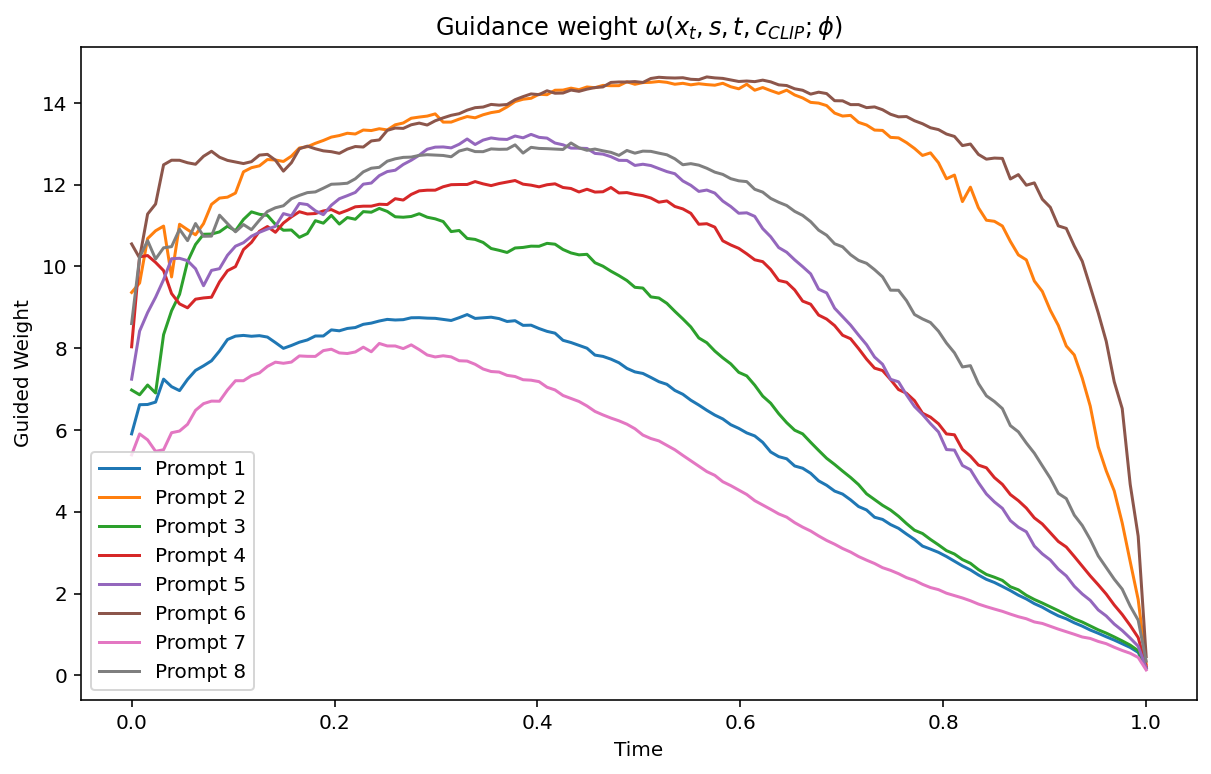}
    \end{minipage}\hfill
    \begin{minipage}[c]{0.54\textwidth}
    \scriptsize 
    \textbf{Prompt 1}. A black dog lays his head on a book of maps. \\
    \textbf{Prompt 2}. A side by side comparison of the same room in the past and present. \\
    \textbf{Prompt 3}. People are cross country skiing through a field. \\
    \textbf{Prompt 4}. A person holds a flip phone displaying the screen. \\
    \textbf{Prompt 5}. A cake has had a slice taken out of it and put onto the plate. \\
    \textbf{Prompt 6}. A display case in front of a store filled with umbrellas. \\
    \textbf{Prompt 7}. A black and white dog laying on a bed. \\
    \textbf{Prompt 8}. A kitchen that has various paintings on the walls and a large fish aquarium against the wall.
    \end{minipage}
    \vspace{-0.1in}
    \caption{\textbf{Learned guidance weights on MS COCO $512\times512$} trained with \Cref{alg:flowmatching-gan-dre} on MMDiT-S.
    Please refer to \Cref{fig:large-ms-coco-visualization-main,fig:large-ms-coco-visualization-additional} for the corresponding images.
    }
    \label{fig:guidance_weights_coco_mmdit_large}
\end{figure}

\begin{figure}[h]
\centering
\begin{subfigure}{\textwidth}
    \includegraphics[width=0.19\textwidth]{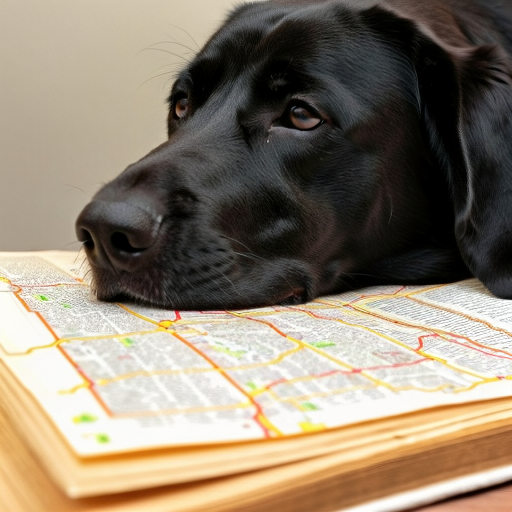}
    \includegraphics[width=0.19\textwidth]{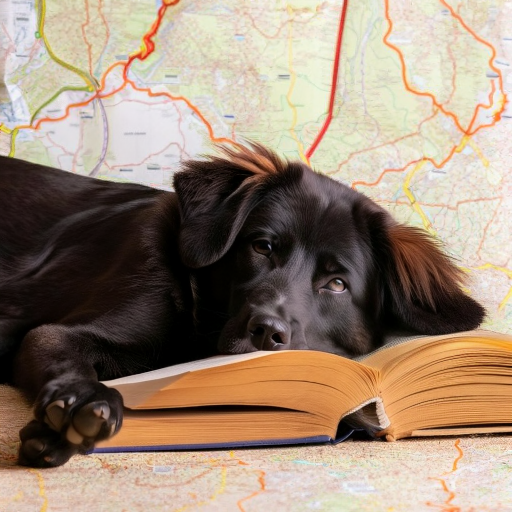}
    \includegraphics[width=0.19\textwidth]{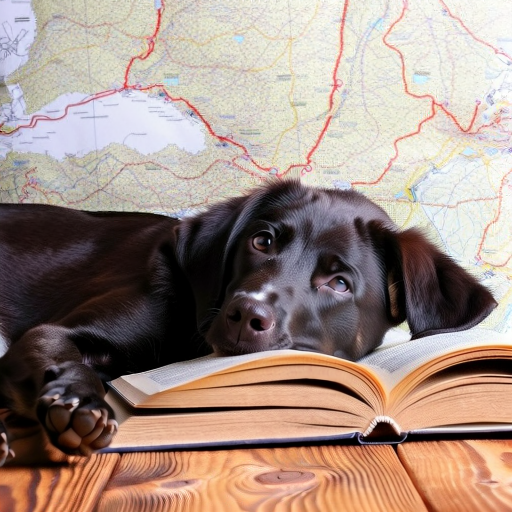}
    \includegraphics[width=0.19\textwidth]{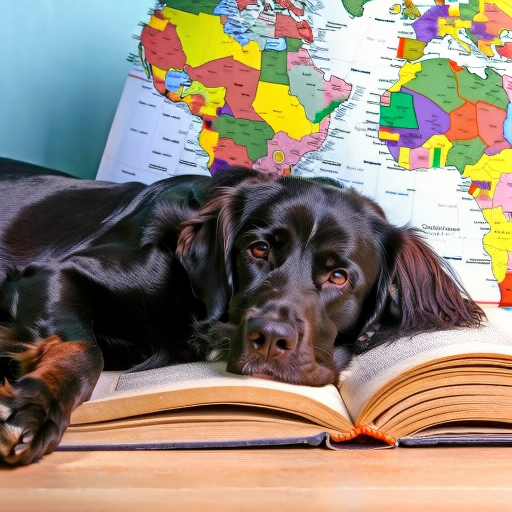}
    \includegraphics[width=0.19\textwidth]{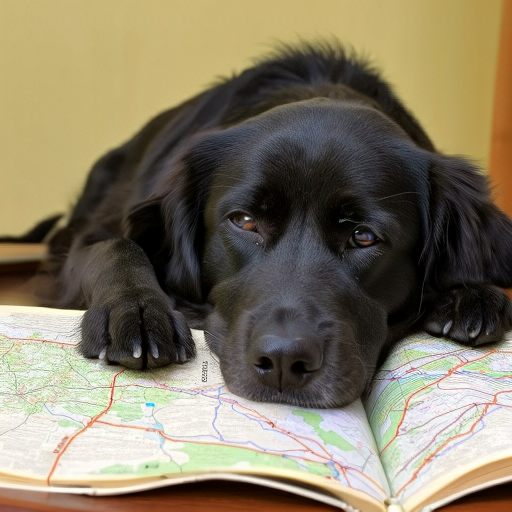}
    \caption{\textit{A black dog lays his head on a book of maps.}}
\end{subfigure}
\hfill
\begin{subfigure}{\textwidth}
    \includegraphics[width=0.19\textwidth]{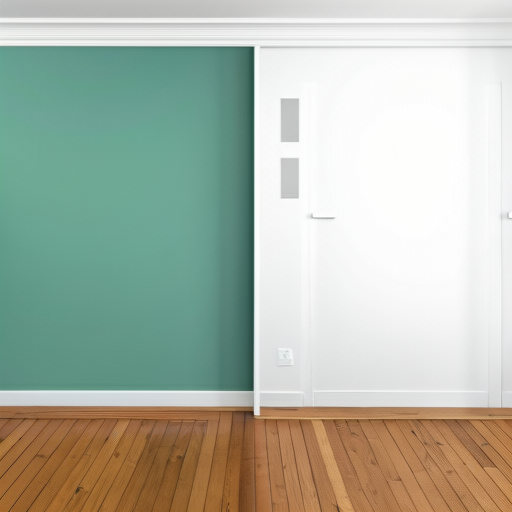}
    \includegraphics[width=0.19\textwidth]{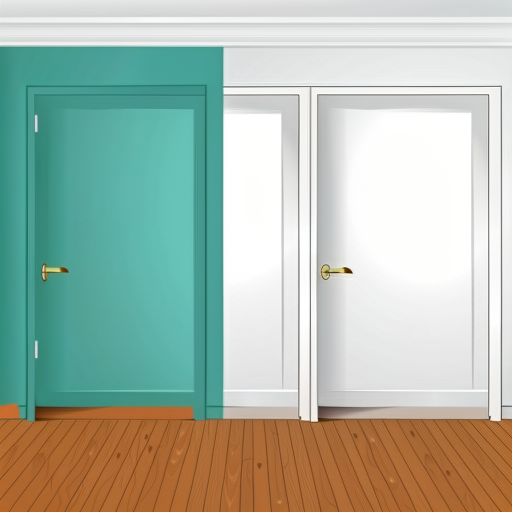}
    \includegraphics[width=0.19\textwidth]{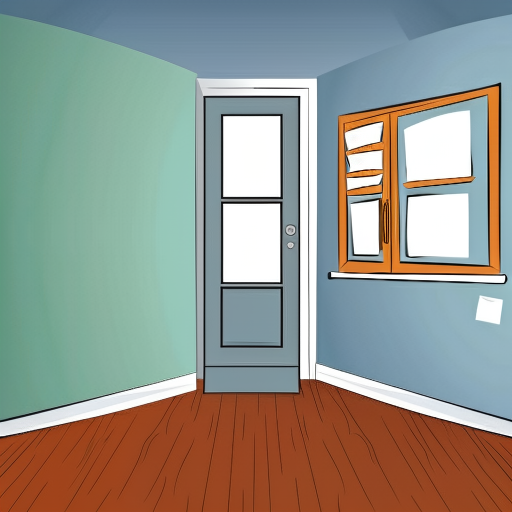}
    \includegraphics[width=0.19\textwidth]{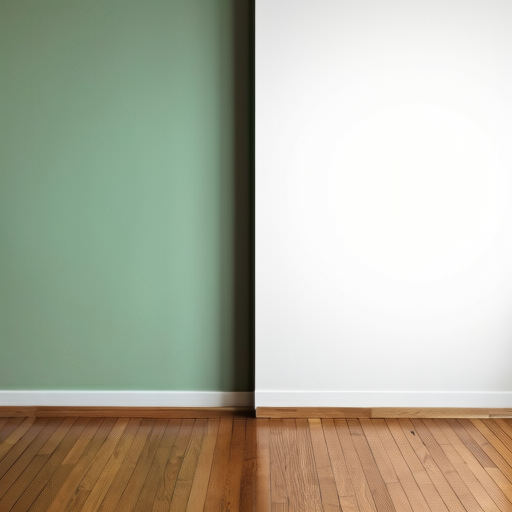}
    \includegraphics[width=0.19\textwidth]{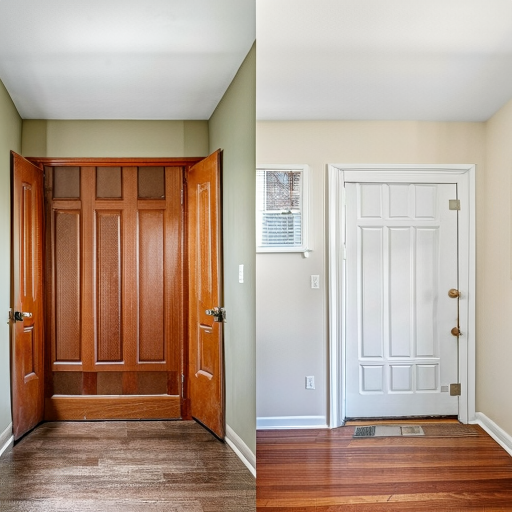}
    \caption{\textit{A side by side comparison of the same room in the past and present.}}
\end{subfigure}
\hfill
\begin{subfigure}{\textwidth}
    \includegraphics[width=0.19\textwidth]{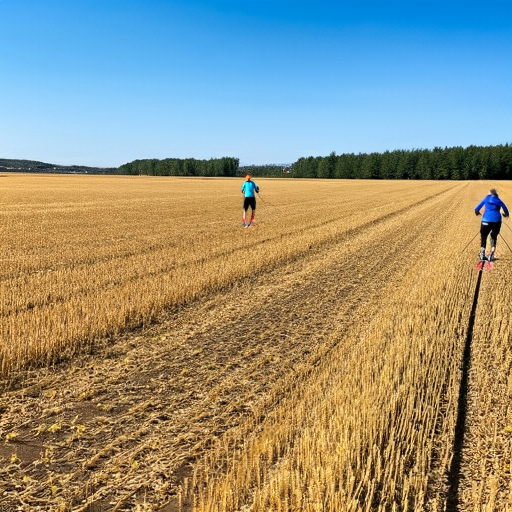}
    \includegraphics[width=0.19\textwidth]{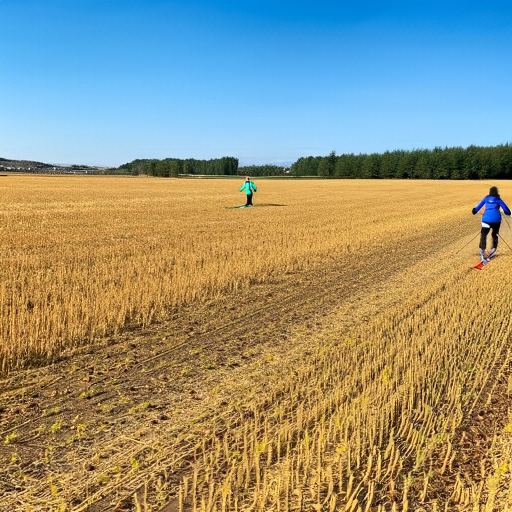}
    \includegraphics[width=0.19\textwidth]{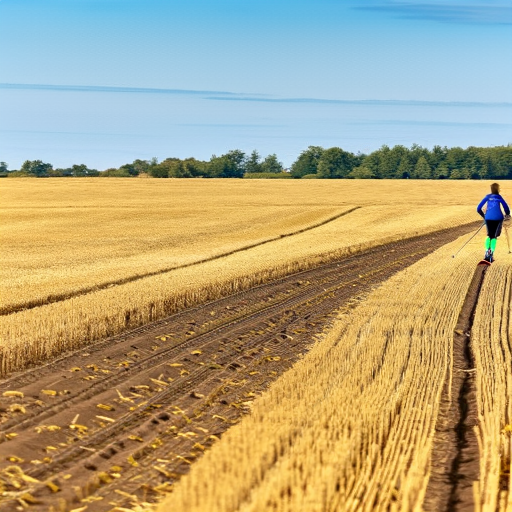}
    \includegraphics[width=0.19\textwidth]{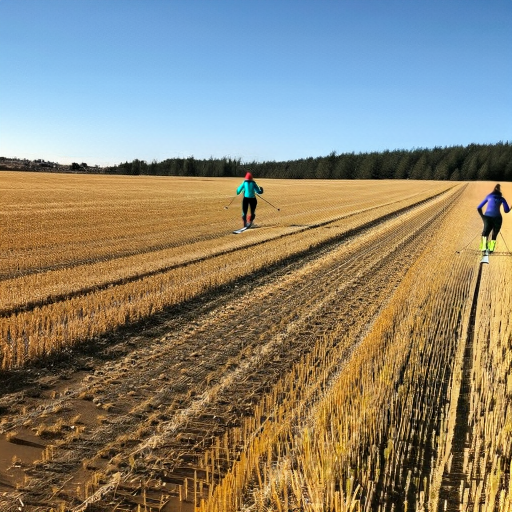}
    \includegraphics[width=0.19\textwidth]{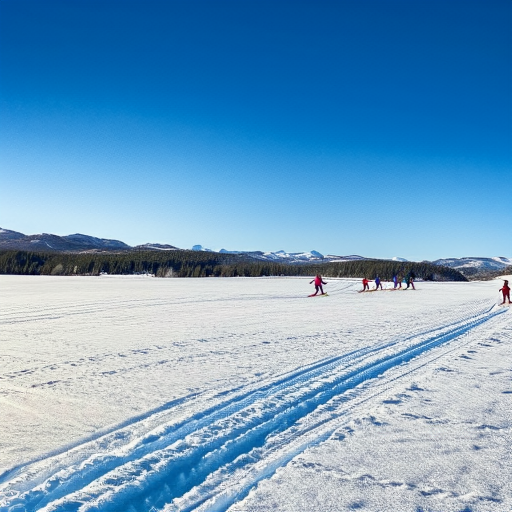}
    \caption{\textit{People are cross country skiing through a field.}}
\end{subfigure}
\caption{\textbf{T2I Results on MS-COCO with MMDiT-S}. (left-to-right) We provide results of images generated from the given text prompt with CFG $\omega = 7.5$, Clamp linear guidance (CLG), Limited Interval guidance (LIG), MMD guidance, and our method. Qualitatively, our method (rightmost) exhibits better adherence to text prompts, precisely capturing nuanced details such as the dog’s interaction with a book of maps in (a), the temporal dichotomy of the split room in (b), and the specific winter-field context in (c) that baselines simplify or omit.}
\label{fig:large-ms-coco-visualization-main}
\end{figure}

\begin{figure}[h]
\centering
\begin{subfigure}{\textwidth}
    \includegraphics[width=0.19\textwidth]{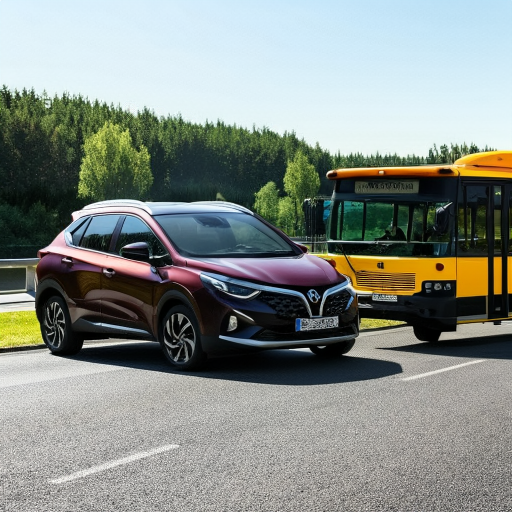}
    \includegraphics[width=0.19\textwidth]{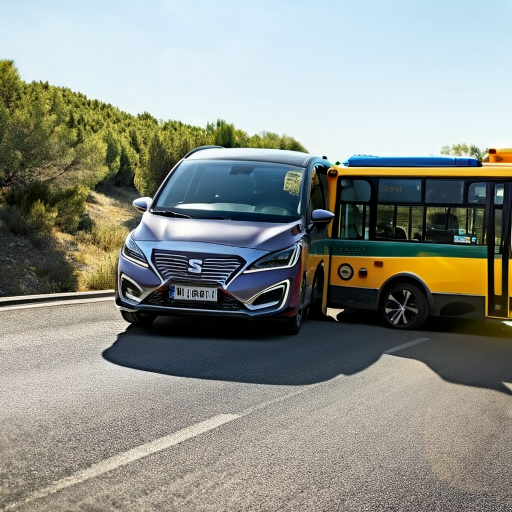}
    \includegraphics[width=0.19\textwidth]{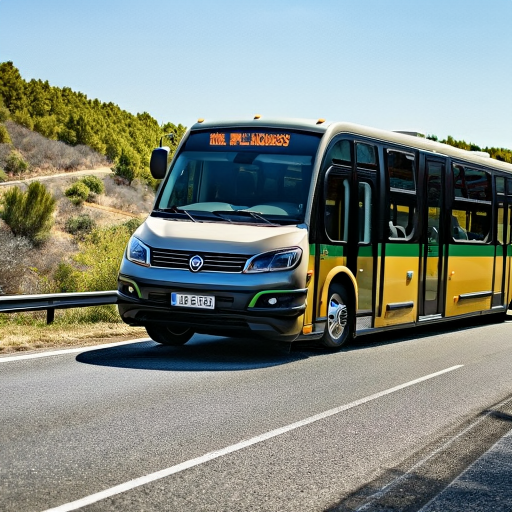}
    \includegraphics[width=0.19\textwidth]{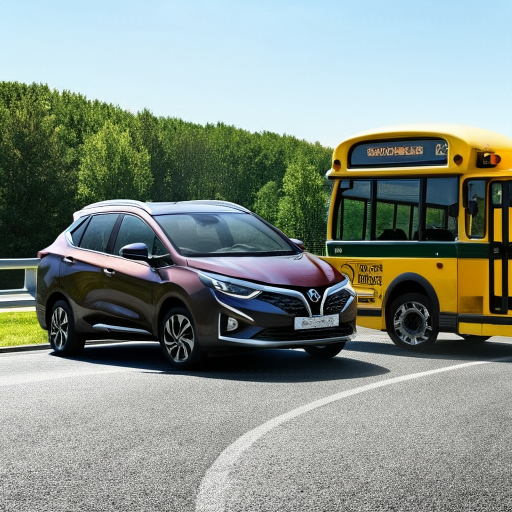}
    \includegraphics[width=0.19\textwidth]{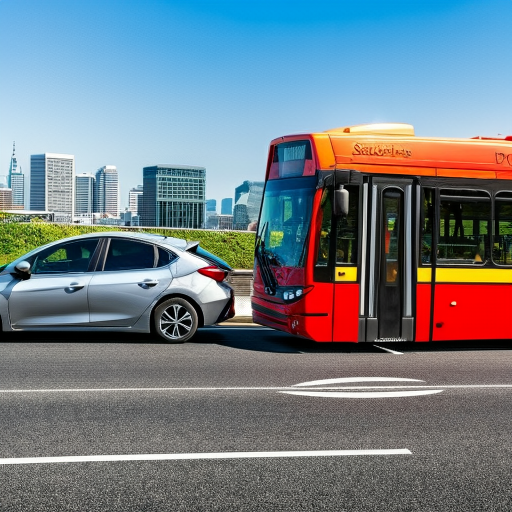}
    \caption{\textit{A car and a public transit vehicle on a road.}}
\end{subfigure}
\hfill
\begin{subfigure}{\textwidth}
    \includegraphics[width=0.19\textwidth]{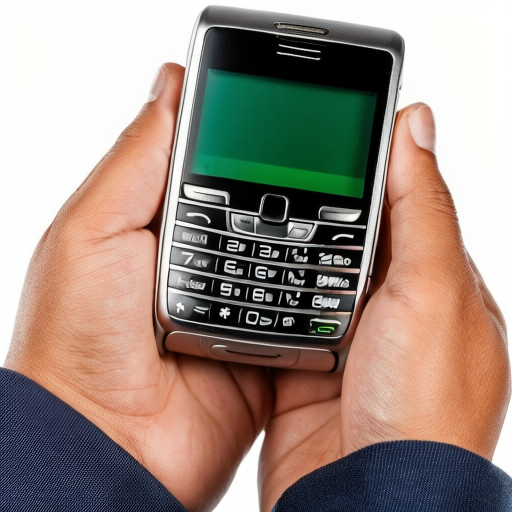}
    \includegraphics[width=0.19\textwidth]{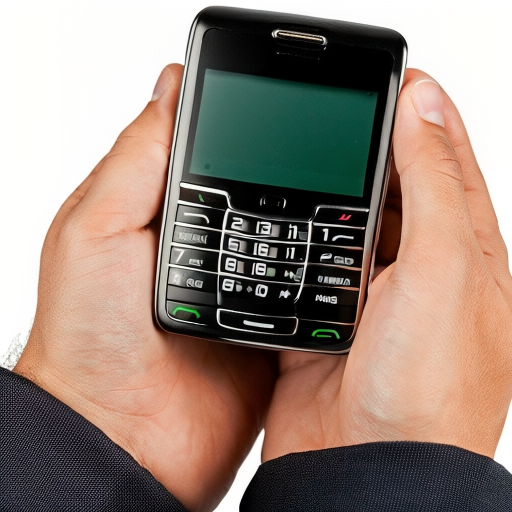}
    \includegraphics[width=0.19\textwidth]{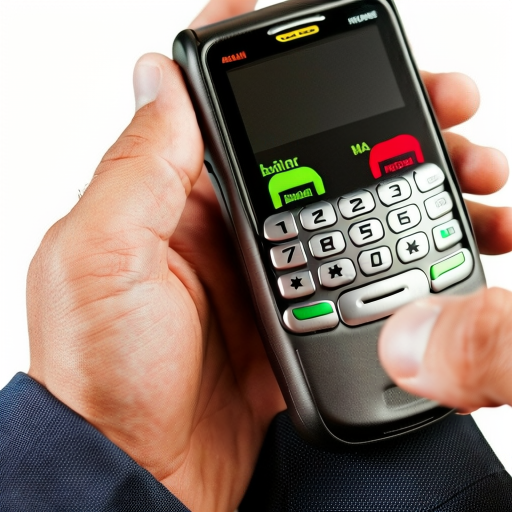}
    \includegraphics[width=0.19\textwidth]{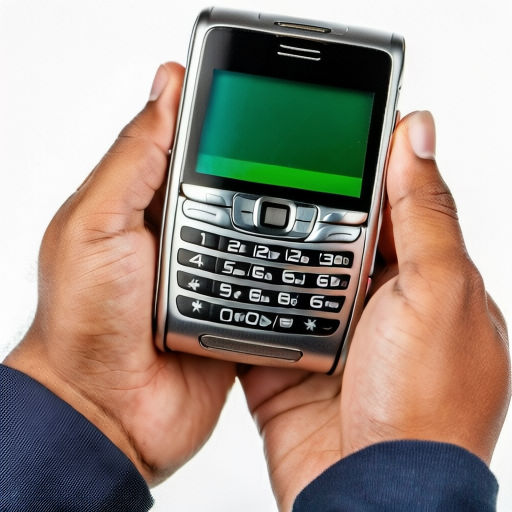}
    \includegraphics[width=0.19\textwidth]{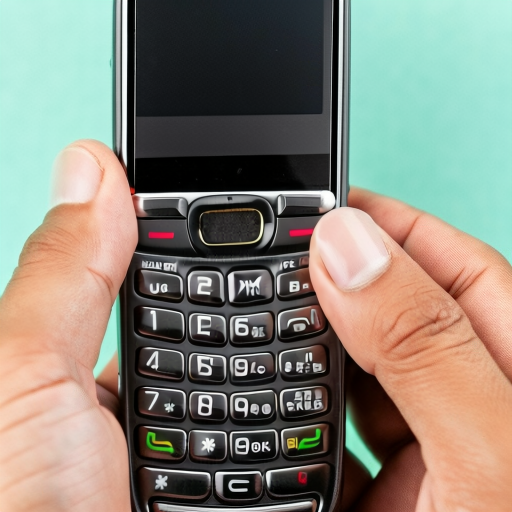}
    \caption{\textit{A person holds a flip phone displaying the screen.}}
\end{subfigure}
\hfill
\begin{subfigure}{\textwidth}
    \includegraphics[width=0.19\textwidth]{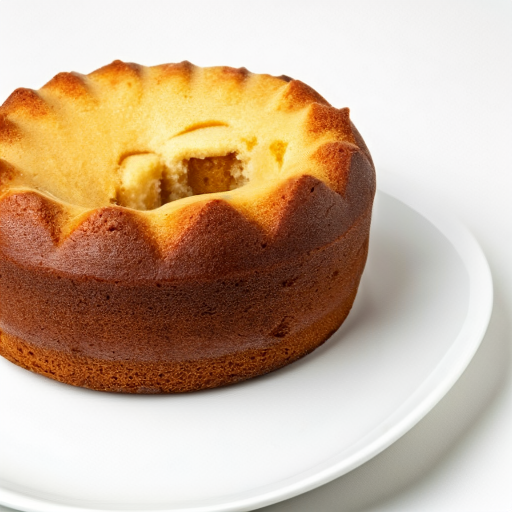}
    \includegraphics[width=0.19\textwidth]{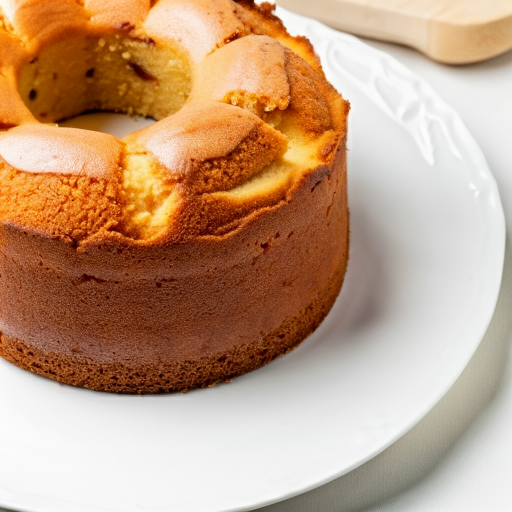}
    \includegraphics[width=0.19\textwidth]{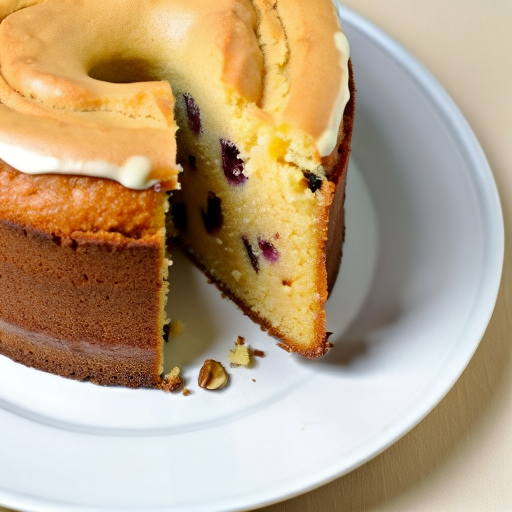}
    \includegraphics[width=0.19\textwidth]{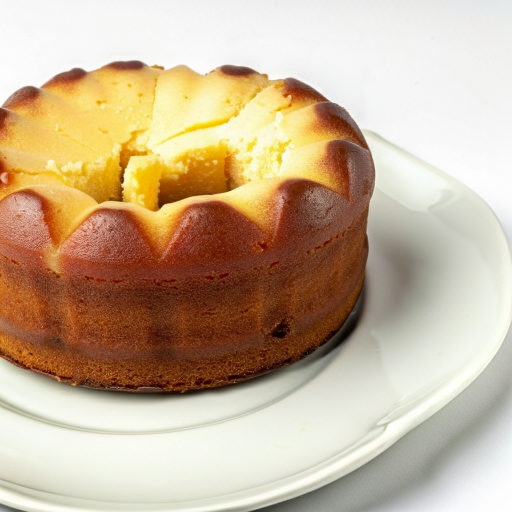}
    \includegraphics[width=0.19\textwidth]{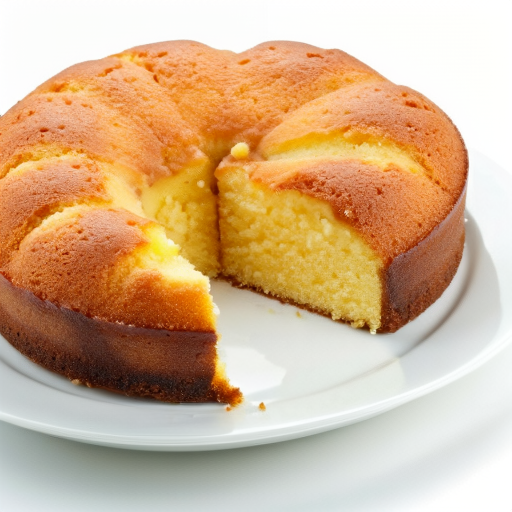}
    \caption{\textit{A cake has had a slice taken out of it and put onto the plate.}}
\end{subfigure}
\hfill
\begin{subfigure}{\textwidth}
    \includegraphics[width=0.19\textwidth]{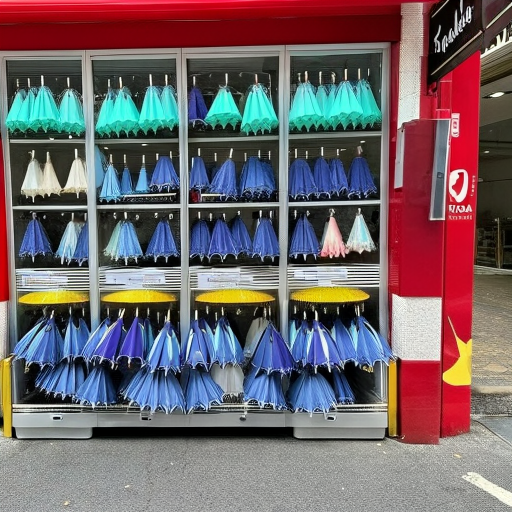}
    \includegraphics[width=0.19\textwidth]{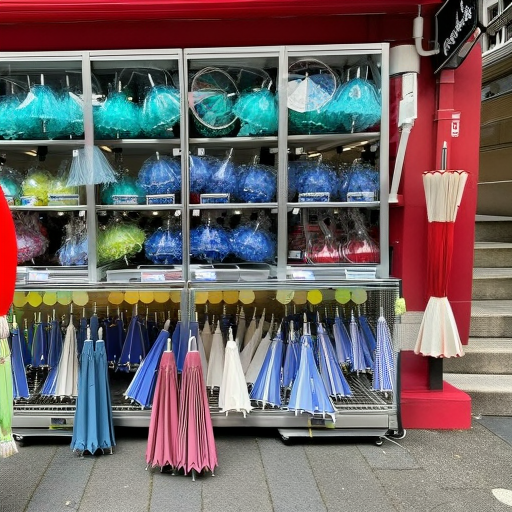}
    \includegraphics[width=0.19\textwidth]{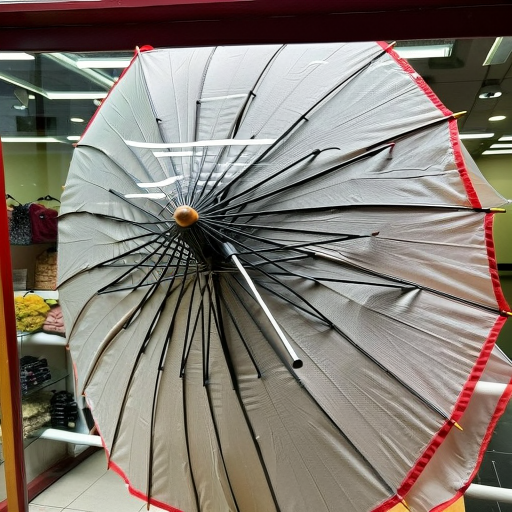}
    \includegraphics[width=0.19\textwidth]{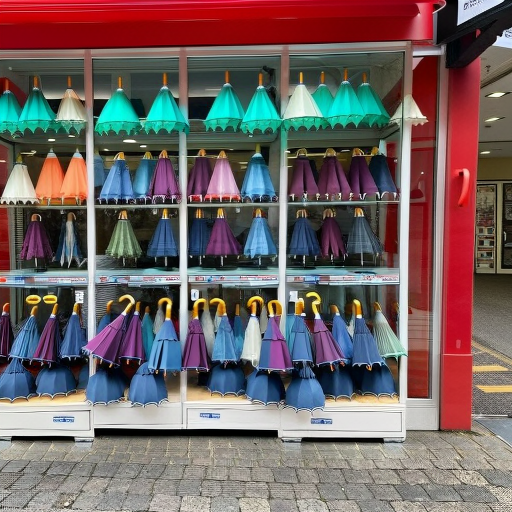}
    \includegraphics[width=0.19\textwidth]{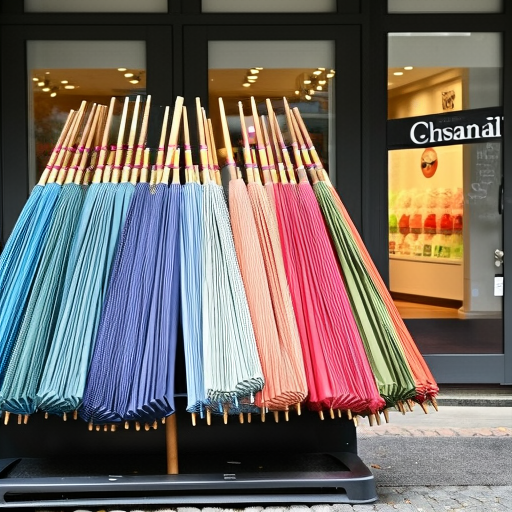}
    \caption{\textit{A display case in front of a store filled with umbrellas.}}
\end{subfigure}
\hfill
\begin{subfigure}{\textwidth}
    \includegraphics[width=0.19\textwidth]{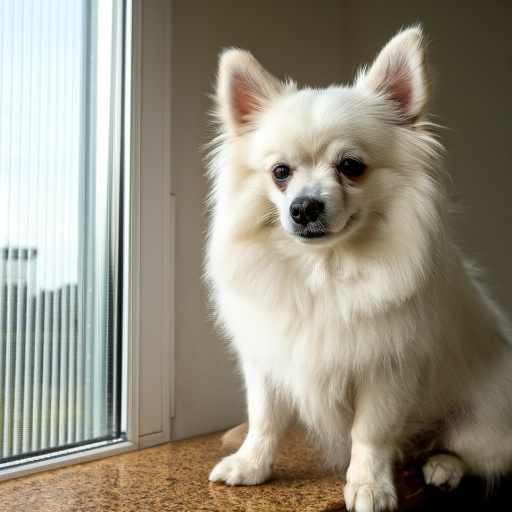}
    \includegraphics[width=0.19\textwidth]{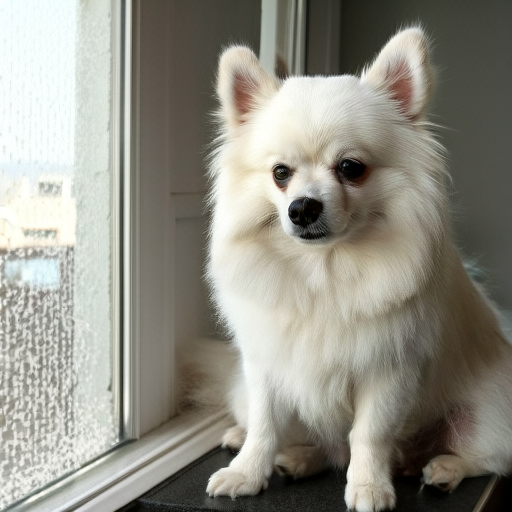}
    \includegraphics[width=0.19\textwidth]{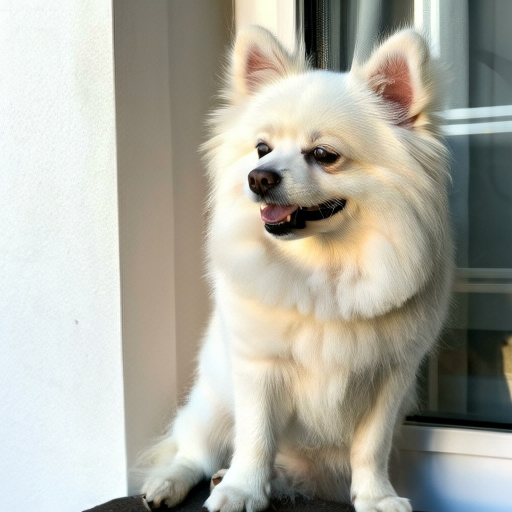}
    \includegraphics[width=0.19\textwidth]{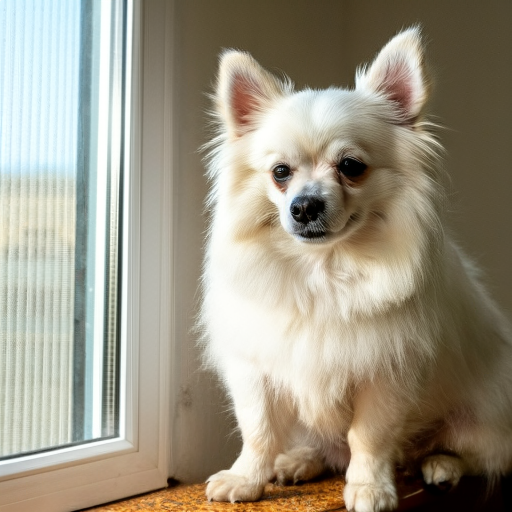}
    \includegraphics[width=0.19\textwidth]{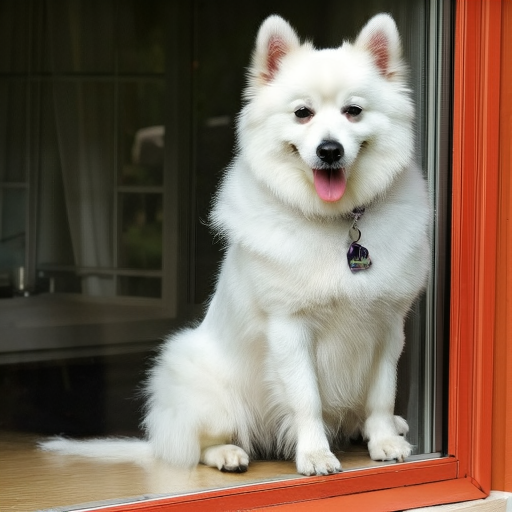}
    \caption{\textit{A white dog sitting on a ledge of a window.}}
\end{subfigure}
\hfill
\begin{subfigure}{\textwidth}
    \includegraphics[width=0.19\textwidth]{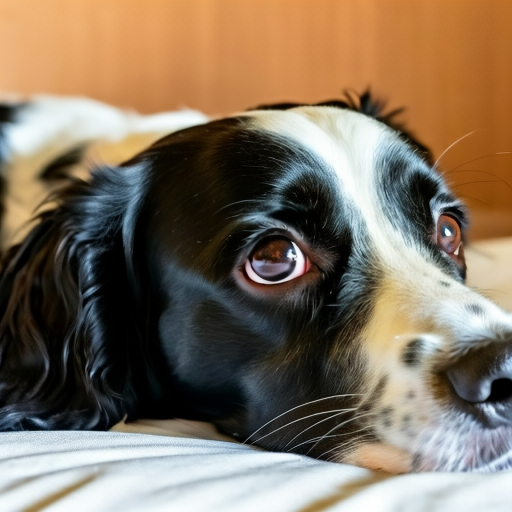}
    \includegraphics[width=0.19\textwidth]{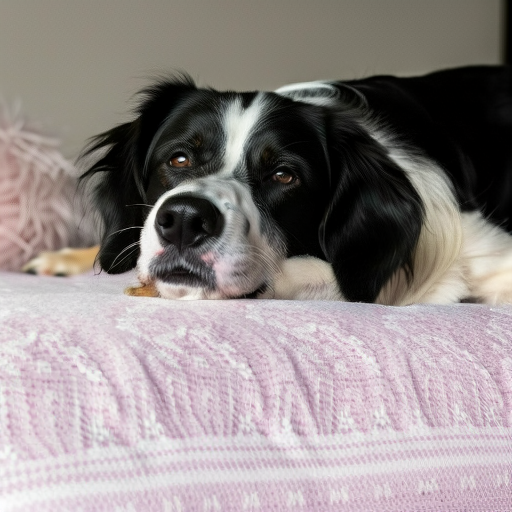}
    \includegraphics[width=0.19\textwidth]{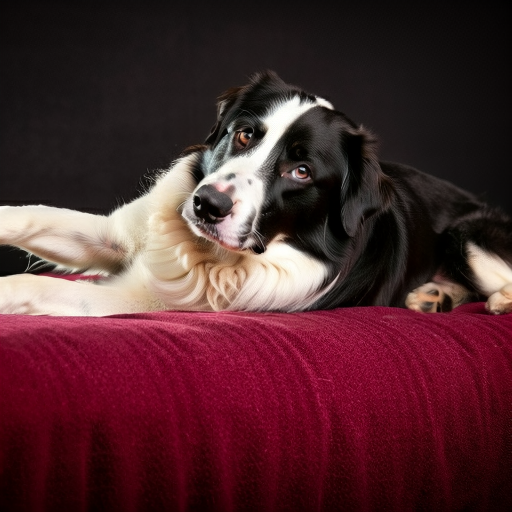}
    \includegraphics[width=0.19\textwidth]{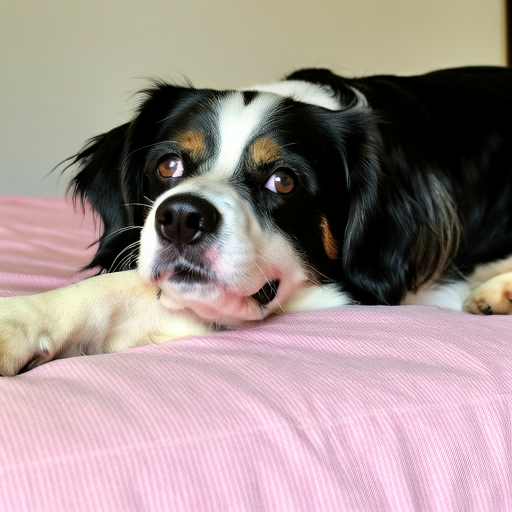}
    \includegraphics[width=0.19\textwidth]{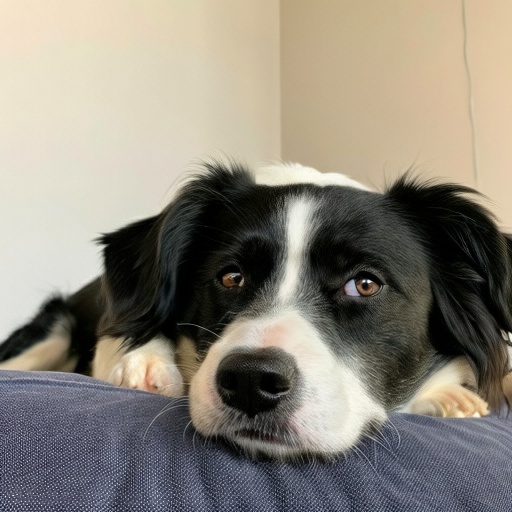}
    \caption{\textit{A black and white dog laying on a bed.}}
\end{subfigure}
\caption{\textbf{T2I Results on MS-COCO with MMDiT-S}. (left-to-right) We provide results of images generated from the given text prompt with CFG $\omega = 7.5$, CLG, LIG, MMD and our method.}
\label{fig:large-ms-coco-visualization-additional}
\end{figure}

\begin{figure}[h]
\centering
\begin{subfigure}{\textwidth}
    \includegraphics[width=0.19\textwidth]{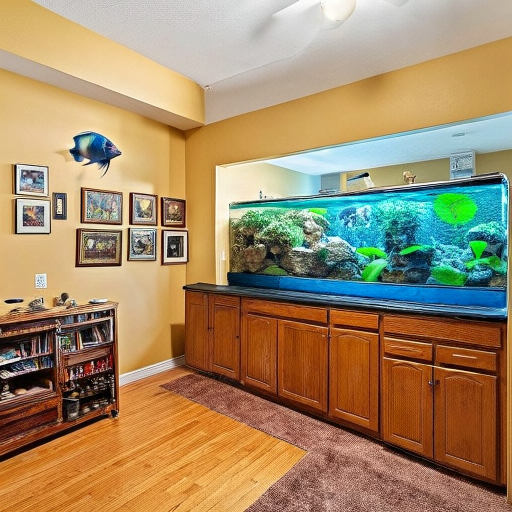}
    \includegraphics[width=0.19\textwidth]{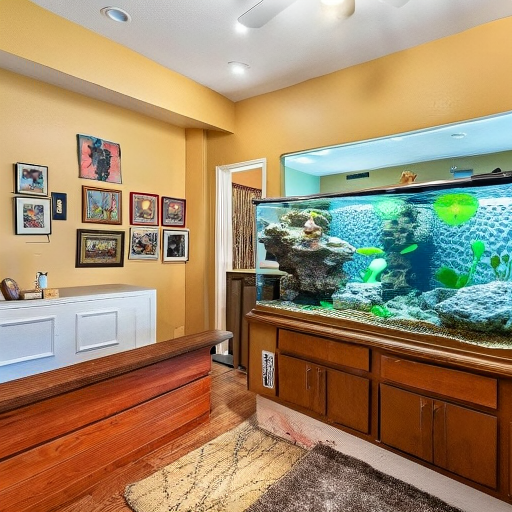}
    \includegraphics[width=0.19\textwidth]{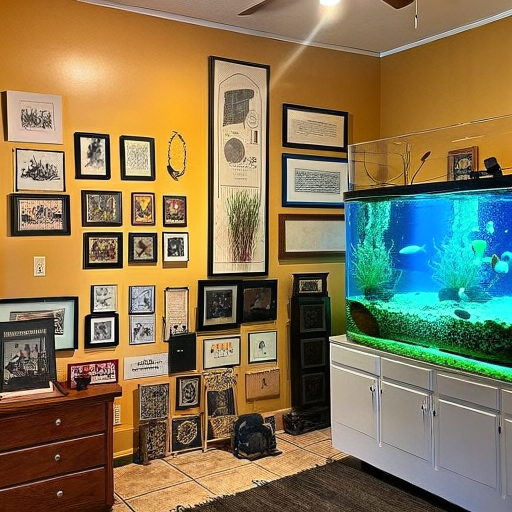}
    \includegraphics[width=0.19\textwidth]{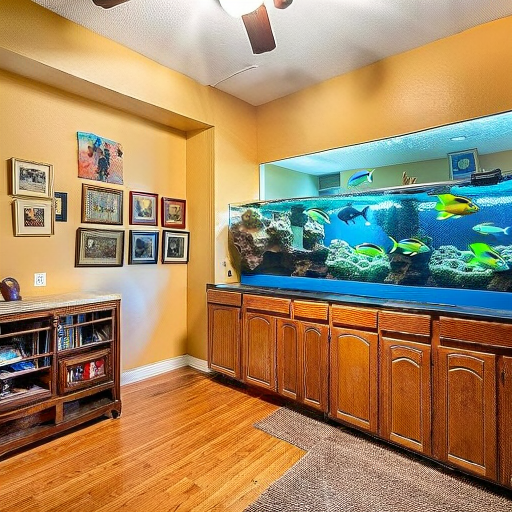}
    \includegraphics[width=0.19\textwidth]{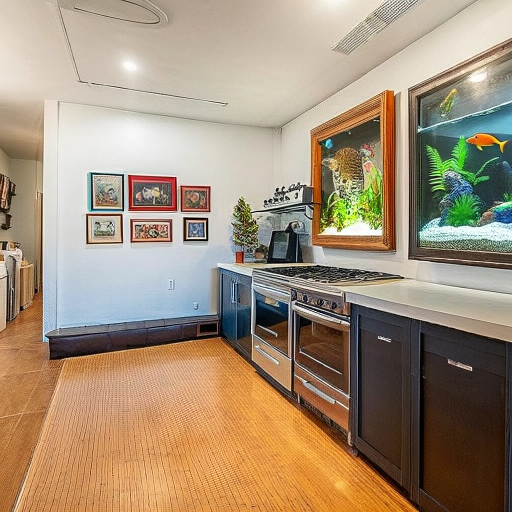}
    \caption{\textit{A kitchen that has various paintings on the walls and a large fish aquarium against the wall.}}
\end{subfigure}
\hfill
\begin{subfigure}{\textwidth}
    \includegraphics[width=0.19\textwidth]{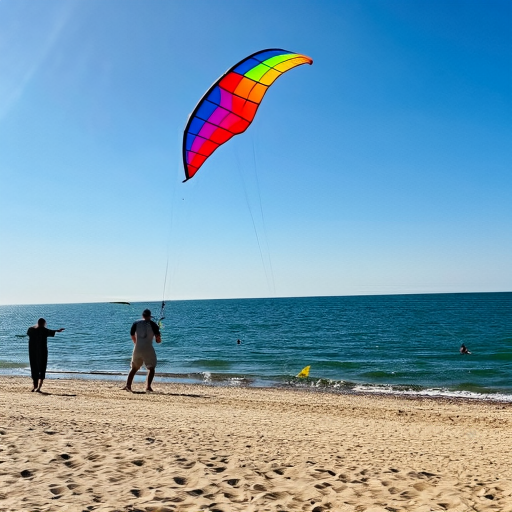}
    \includegraphics[width=0.19\textwidth]{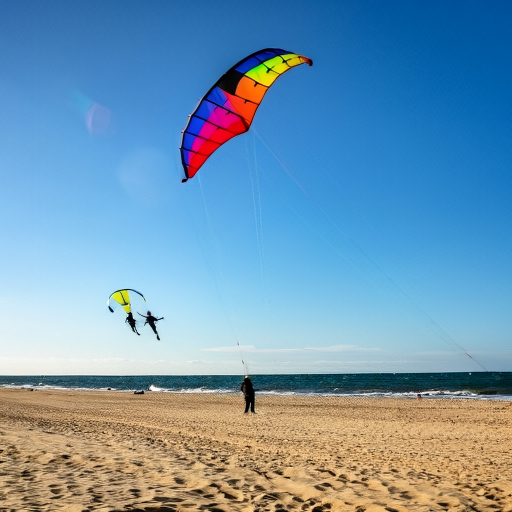}
    \includegraphics[width=0.19\textwidth]{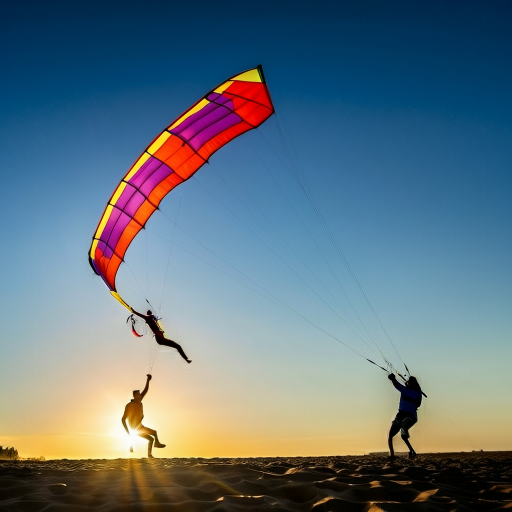}
    \includegraphics[width=0.19\textwidth]{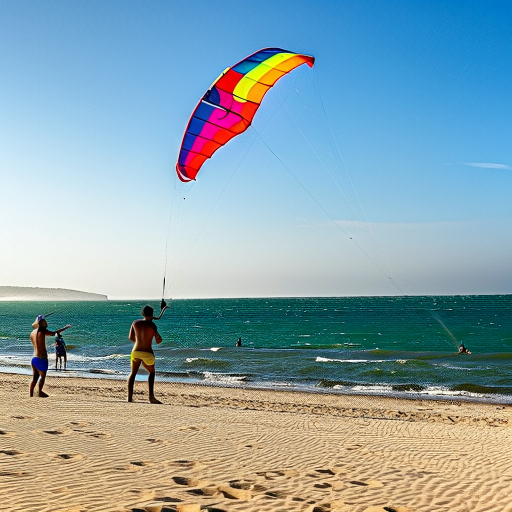}
    \includegraphics[width=0.19\textwidth]{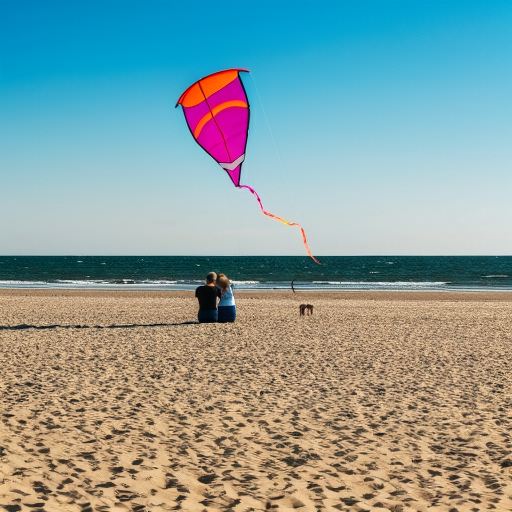}
    \caption{\textit{A couple of people are flying a kite on top of a sandy beach.}}
\end{subfigure}
\caption{\textbf{T2I Results on MS-COCO with MMDiT-S}. (left-to-right) We provide results of images generated from the given text prompt with CFG $\omega = 7.5$, CLG, LIG, MMD and our method.}
\label{fig:large-ms-coco-visualization-additional-2}
\end{figure}

We provide qualitative results in \Cref{fig:small-ms-coco-visualization-additional} for MMDiT-XS, and in \Cref{fig:large-ms-coco-visualization-main,fig:large-ms-coco-visualization-additional,fig:large-ms-coco-visualization-additional-2} for MMDiT-S. 
Our method generates images that are more realistic and better aligned with the text prompts. The learned guidance weights are shown in~\Cref{fig:guidance_weights_coco_mmdit_large}. 
We observe high variability depending on the prompt.

\subsection{Ablations}

\textbf{Conditioning Information.} We run the ablation on the kind of information the guidance network $\boldsymbol{\omega}$ receives.
Previously, ~\citet{galashov2025learn} concluded that passing both time-steps $(t,s)$ and conditioning $c$ was strictly better than less information. In~\Cref{tab:mmdit_cond_information}, we show that adding $x_t$ to the guidance network, consistently improves performance but hurts FID.

\begin{table}[h]
\centering
\caption{\textbf{MS COCO $512 \times 512$}. Impact of conditioning information on empirical performance of our method. In \textbf{bold} we highlight the best metric among each model size.}
\begin{tabular}{lccccc}
\toprule
\textbf{Conditioning} & \textbf{FID $\downarrow$} & \textbf{CLIP $\uparrow$} & \textbf{Aesthetic $\uparrow$} & \textbf{HPSv2 $\uparrow$} & \textbf{PickScore $\uparrow$} \\
\midrule
MMDiT-XS, $\boldsymbol{\omega}(s, t, c)$     & \textbf{26.91} & 0.30368 & 5.27 & 0.28112 & 0.21778 \\
MMDiT-XS, $\boldsymbol{\omega}(x_s,s, t, c)$ &
31.31 & \textbf{0.30483} & \textbf{5.32} & \textbf{0.2829} & \textbf{0.2186} \\
\midrule
MMDiT-S, $\boldsymbol{\omega}(s, t, c)$      & \textbf{28} & 0.3063 & 5.28 & 0.2841 & 0.2206 \\
MMDiT-S, $\boldsymbol{\omega}(x_s,s, t, c)$  & 31.7 & \textbf{0.3068} & \textbf{5.34} & \textbf{0.28564} & \textbf{0.2214} \\
\bottomrule
\end{tabular}
\label{tab:mmdit_cond_information}
\end{table}

\textbf{Time distribution.} We study impact of parameter $\delta$ on performance of MMDiT-XS. The results are given in~\Cref{tab:mmdit_delta_study}. We see that overall the performance is quite robust to the choice of $\delta$. We also provide comparisons to alternative time distribution $p(s,t)$ in~\Cref{sec_app:other_time_dist}.

\textbf{Additional ablations.} Due to space contraints, we provide additional ablations in the Appendix sections. In~\Cref{sec_app:number_of_sampling_steps}, we study the impact of sampling steps on the performance of our method. On top of that, in~\Cref{sec_app:ablation_gen_arch}, we study alternative choice of the generator network architecture. Finally, in~\Cref{sec_app:noise_impact}, we study the impact of noise in training and in~\Cref{sec_app:noise_ablation} we study the impact of how the pairs of $(x_1,c)$ are sampled.

\begin{table}[h]
\centering
\caption{Impact of $\delta$ variations on MMDiT-XS performance.}
\begin{tabular}{cccccc}
\toprule
$\bm{\delta}$ & \textbf{FID $\downarrow$} & \textbf{CLIP $\uparrow$} & \textbf{Aesthetic $\uparrow$} & \textbf{HPSv2 $\uparrow$} & \textbf{PickScore $\uparrow$} \\
\midrule
$0.01$ & 31.2 & 0.30476 & 5.32 & 0.28273 & 0.21888 \\
$0.10$  & 31.31 & 0.30484 & 5.32 & 0.28288 & 0.21862 \\
$0.20$  & 30.3 & 0.30486 & 5.31 & 0.28263 & 0.21850 \\
\bottomrule
\end{tabular}
\label{tab:mmdit_delta_study}
\end{table}

\section{Discussion and Limitations}
We introduce an adversarial method to learn dynamic guidance schedules by enforcing a marginal consistency condition along the diffusion sampling trajectory. 
We train a lightweight MLP through an objective that matches the marginal densities of the guided distribution and true data distribution along the sampling trajectory.
Empirically, our method outperforms constant CFG schedule, heuristic-based dynamic  schedules, as well as, prior learned guidance approaches on human preference and text alignment metrics on the MS-COCO $512\times512$ benchmark.
Despite its empirical success, our method has several limitations that provide avenues for future work.
First, while our learned schedule introduces negligible computational overhead during inference, the training procedure is complex.
Optimizing an adversarial game over a diffusion trajectory requires careful choice of training objectives, optimization strategies such as Two-Time-Scale Update Rule (TTUR). 
Second, the learned guidance scheduler is backbone-specific. As the network relies on the local geometric statistics of a specific pre-trained flow matching model (e.g., MMDiT), the scheduler cannot be transferred zero-shot to a different architecture or velocity field without re-training.
Finally, our objective for T2I incorporates a CLIP reward loss. There are avenues to explore alternate reward signals that might overcome drawbacks of CLIP reward~\citep{kang2025clip}, which we leave as future work.

\bibliography{references}
\bibliographystyle{iclr2027_conference}

\newpage

\appendix

\section*{Organization of the appendix}

In~\cref{sec_app:ext_rel_work}, we present the extended related work. In~\cref{sec:baseline_ablations}, we present additional results and ablations which were not included in the main paper. It includes additional experimental comparisons to baselines in Appendix~\ref{sec:baseline_ablations}. Moreover, it includes additional ablations over the hyperparameters for the considered baselines, see Appendix~\ref{sec_app:t2i_baseline_ablations}. Appendix~\ref{sec_app:time_dist_ablations_main} shows results for the parameters of the time distribution $p(s,t)$ from~\Cref{sec:guidance_learning_algo}. On top of that, in Appendix~\ref{sec_app:number_of_sampling_steps}, we show how the number of sampling steps affects performance of our guidance method. Moreover, in Appendix~\ref{sec_app:ablation_gen_arch}, we show the impact of the generator network architecture for learning the guidance weights. Furthermore, in Appendix~\ref{sec_app:noise_ablation}, we study the importance of the correlation in the noise distribution of clean data for the Algorithm~\ref{alg:flowmatching-gan-dre}. In Appendix~\ref{sec_app:how_image_pairs_sampled}, we study the impact of how the pairs $(x_1,c)$ are sampled. Finally, in Appendix~\ref{sec_app:other_time_dist} we discuss alternative choices for the time distribution. In Appendix~\ref{app_sec:experimental_details}, we provide the experimental details and in Appendix~\ref{sec_app:image_samples}, we show different image samples from various models.

\section{Extended Related Work}
\label{sec_app:ext_rel_work}

\textbf{Approaches for General Guidance.} Beyond the scope of CFG, a broader class of methods aim to guide diffusion model by leveraging external models or constraints without requiring any re-training. Universal guidance~\citep{bansal2023universal} pioneered this area by demonstrating that the gradients of any differentiable model such as the CLIP feature extractor~\citep{radford2021learning}, a segmentation network, and an object detection network, can be used guide the diffusion sampling process towards specific attributes.
Loss-based guidance methods such as FreeDoM~\citep{yu2023freedom} and subsequent works~\citep{shen2024understanding} treat the guidance process as a form of energy-based optimization where the latent state is iteratively udpated to minimize a user-defined loss.

\textbf{Distillation.}
Our approach also shares conceptual similarities with the literature on diffusion distillation, especially adversarial distillation methods.
Distribution matching distillation (DMD, DMDv2)~\citep{yin2024one,yin2024improved} and latent adversarial diffusion distillation (LADD)~\citep{sauer2024fast} utilize a discriminator to match the marginal distribution of generated samples to the real data distribution, eliminating the blurriness from trajectory-based distillation methods such as Progressive distillation~\citep{salimans2022progressive} and consistency models~\citep{song2023consistency}. 
$f$-distill~\citep{xu2025one} further generalizes this by minimizing $f$-divergence via an adversarial game.

\section[]{Additional results and ablations}
\label{sec:baseline_ablations}

\subsection{Extensive comparisons to baselines}
\label{sec:extensive_comparisons}

To complement the results from~\Cref{sec:t2i_generation}, we add additional comparison to~\citep{galashov2025learn} baseline in \Cref{table:ms-coco-results-small-full} and in \Cref{table:ms-coco-results-large-full}.

\begin{table}[htbp!]
\centering
\caption{\textbf{MS COCO $512 \times 512$}. Performance of different methods on a small, MMDiT-XS model. We highlight the best metric in \textbf{bold}. $\dagger$ indicates that we follow the setup from \citet{galashov2025learn}.
}
\resizebox{\textwidth}{!}{
\begin{tabular}{lcccccc}
\toprule
\textbf{Method} & \textbf{Guidance Weight} & \textbf{FID $\downarrow$} & \textbf{CLIP $\uparrow$} & \textbf{Aesthetic $\uparrow$} & \textbf{HPSv2 $\uparrow$} & \textbf{PickScore $\uparrow$} \\
\midrule
& \textbf{Baselines} \\
\midrule
Unguided & $\omega = 0$ & 27.65 & 0.2742 & 4.79 & 0.2547 & 0.1997 \\
Constant & $\omega = 7.5$ & 29.73 & 0.3044 & 5.29 & 0.2817 & 0.2186 \\
LIG& $\omega(t) = 12.5 \in [0.1, 0.9]$ & \textbf{25.57} & 0.3032 & 5.30 & 0.2812 & 0.2171 \\ %
CLG & $\omega = 15.0$ & 26.38 & 0.3038 & 5.26 & 0.2807 & 0.2175 \\
MMD + SC & $\boldsymbol{\omega}(s,t,c)^\dagger$ & 28.39 & 0.30388 & 5.25 & 0.28 & 0.21795 \\
MMD + SC & $\boldsymbol{\omega}(s,t,c)$ & 27.72132 & 0.30385 & 5.25978  & 0.28101 & 0.21792 \\
MMD + SC & $\boldsymbol{\omega}(x_s,s,t,c)$ & 28.38 & 0.30419 & 5.2647 & 0.28129 & 0.21813 \\
MMD + MC & $\boldsymbol{\omega}(s,t,c)^\dagger$ & 26.37662 & 0.30344 & 5.23274 & 0.27991 & 0.21733 \\
MMD + MC & $\boldsymbol{\omega}(s,t,c)$ & 26.21882 & 0.30369 & 5.23151 & 0.27992 & 0.21725 \\
MMD + MC & $\boldsymbol{\omega}(x_s,s,t,c)$ & 28.10983 & 0.30368 & 5.25155 & 0.28074 & 0.21789 \\
\midrule
GAN + MC & $\boldsymbol{\omega}(x_s,s,t,c)$ & 31.31 & \textbf{0.3048} & \textbf{5.32} & \textbf{0.2829} & \textbf{0.2186} \\
\bottomrule
\end{tabular}
}
\label{table:ms-coco-results-small-full}
\end{table}

\begin{table}[htbp!]
\centering
\caption{\textbf{MS COCO $512 \times 512$}. Performance of different methods on a large, MMDiT-S model. We highlight the best metric in \textbf{bold}. $\dagger$ indicates that we follow the setup from \cite{galashov2025learn}.
}
\resizebox{\textwidth}{!}{
\begin{tabular}{ccccccc}
\toprule
\textbf{Method} & \textbf{Guidance Weight} & \textbf{FID $\downarrow$} & \textbf{CLIP $\uparrow$} & \textbf{Aesthetic $\uparrow$} & \textbf{HPSv2 $\uparrow$} & \textbf{PickScore $\uparrow$} \\
\midrule
& \textbf{Baselines} \\
\midrule
Unguided & $\omega = 0$ & 24.3 & 0.279 & 4.84 & 0.2588 & 0.2024 \\
Constant & $\omega = 7.5$ & 31.11 & 0.3063 & 5.31 & 0.285 & \textbf{0.2214} \\
LIG & $\omega(t) = 15.0, t \in [0.1, 0.9]$ & \textbf{24.95} & 0.3042 & 5.31 & 0.2843 & 0.2194 \\
CLG& $\omega = 14.0$ & 27.14 & 0.3055 & 5.27 & 0.2838 & 0.2203 \\
MMD + SC & $\boldsymbol{\omega}(s,t,c)^\dagger$ & 28.5 & 0.30615 & 5.27 & 0.28351 & 0.2204 \\
MMD + SC & $\boldsymbol{\omega}(s,t,c)$ & 28.71683 & 0.30602 & 5.27 & 0.28354 &  0.22049 \\
MMD + SC & $\boldsymbol{\omega}(x_s,s,t,c)$ & 28.54873 & 0.30674 & 5.29430 & 0.28471 & 0.22078 \\
MMD + MC & $\boldsymbol{\omega}(s,t,c)^\dagger$ & 28.8 & 0.3063 & 5.27 & 0.28353 & 0.220 \\
MMD + MC & $\boldsymbol{\omega}(s,t,c)$ & 27.74241 & 0.30622 &  5.25800 & 0.28318 & 0.22016 \\
MMD + MC & $\boldsymbol{\omega}(x_s,s,t,c)$ & 29.43209 & 0.30625 & 5.27623 & 0.28364 & 0.22049 \\
\midrule
GAN + MC & $\boldsymbol{\omega}(x_s,s,t,c)$ & 31.7 & \textbf{0.3068} & \textbf{5.34} & \textbf{0.28564} & \textbf{0.2214} \\
\bottomrule
\end{tabular}
}
\label{table:ms-coco-results-large-full}
\end{table}

\subsection{Text-to-Image Generation baseline ablations}
\label{sec_app:t2i_baseline_ablations}

In~\Cref{table:small-ms-coco-results-ablation-lig-128-steps}, we present results on MMDiT-XS model of limited interval guidance~\citep{kynkaanniemi2024applying} baseline with a large sweep over the hyperparameters such as the guidance weight and the interval. Furthermore, in~\Cref{table:small-ms-coco-results-ablation-clg-128-steps}, we present results for clap-linear guidance schedule~\citep{wang2024analysis}.

\begin{table}[htbp!]
\centering
\caption{\textbf{MS COCO $512 \times 512$}. Performance of limited interval guidance~\citep{kynkaanniemi2024applying} on the MMDiT-XS model with $128$ sampling steps. We highlight the best metrics in \textbf{bold}.}
\begin{tabular}{ccccccc}
\toprule
\textbf{Guidance Weight} & \textbf{Interval} & \textbf{FID $\downarrow$} & \textbf{CLIP $\uparrow$} & \textbf{Aesthetic $\uparrow$} & \textbf{HPSv2 $\uparrow$} & \textbf{Pickscore $\uparrow$} \\
\midrule
$7.5$ & $[0.1, 0.9]$ & 23.97 & 0.3018 & 5.26 & 0.2795 & 0.2166 \\
$7.5$ & $[0.1, 0.8]$ & 24.00 & 0.3018 & 5.26 & 0.2795 & 0.2166 \\
$7.5$ & $[0.2, 0.9]$ & \textbf{19.65} & 0.2969 & 5.17 & 0.2734 & 0.2124 \\
$7.5$ & $[0.2, 0.8]$ & 19.66 & 0.297 & 5.17 & 0.2735 & 0.2124 \\
\midrule
$10.0$ & $[0.1, 0.9]$ & 24.70 & 0.3024 & 5.29 & 0.2805 & 0.217 \\
$10.0$ & $[0.1, 0.8]$ & 24.83 & 0.2977 & 5.29 & 0.2806 & 0.217 \\
$10.0$ & $[0.2, 0.9]$ & 20.32 & 0.3024 & 5.20 & 0.2747 & 0.2129 \\
$10.0$ & $[0.2, 0.8]$ & 20.3 & 0.2976 & 5.21 & 0.2747 & 0.2129 \\
\midrule
$12.5$ & $[0.1, 0.9]$ & 25.57 & \textbf{0.3032} & 5.30 & 0.2812 & \textbf{0.2171} \\
$12.5$ & $[0.1, 0.8]$ & 25.68 & 0.2985 & 5.30 & 0.2812 & \textbf{0.2171} \\
$12.5$ & $[0.2, 0.9]$ & 21.48 & 0.3031 & 5.23 & 0.2755 & 0.2131 \\
$12.5$ & $[0.2, 0.8]$ & 21.57 & 0.2985 & 5.23 & 0.2755 & 0.2132 \\
\midrule
$15.0$ & $[0.1, 0.9]$ & 25.31 & 0.3029 & \textbf{5.31} & \textbf{0.2815} & 0.2170 \\
$15.0$ & $[0.1, 0.8]$ & 25.42 & 0.3029 & \textbf{5.31} & \textbf{0.2815} & \textbf{0.2171} \\
$15.0$ & $[0.2, 0.9]$ & 21.0 & 0.2982 & 5.24 & 0.2760 & 0.2132 \\
$15.0$ & $[0.2, 0.8]$ & 21.0 & 0.2981 & 5.24 & 0.2760 & 0.2132 \\
\bottomrule
\end{tabular}
\label{table:small-ms-coco-results-ablation-lig-128-steps}
\end{table}

\begin{table}[htbp!]
\centering
\caption{\textbf{MS COCO $512 \times 512$}. Performance of clamp-linear guidance schedule $w_t = \text{max}(c_1, \frac{2w_1t} {T})$~\citep{wang2024analysis} on the MMDiT-XS model with $128$ sampling steps. We highlight the best metrics in \textbf{bold}.}
\begin{tabular}{ccccccc}
\toprule
\textbf{Guidance Weight} $\bm{\omega_1}$ & $\bm{c_1}$ & \textbf{FID $\downarrow$} & \textbf{CLIP $\uparrow$} & \textbf{Aesthetic $\uparrow$} & \textbf{HPSv2 $\uparrow$} & \textbf{Pickscore $\uparrow$} \\
\midrule
$7.5$ & $1$ & \textbf{22.37} & 0.3012 & 5.17 & 0.2769 & 0.2167 \\
$7.5$ & $2$ & 23.74 & 0.302 & 5.18 & 0.2778 & 0.2161 \\
$7.5$ & $4$ & 26.82 & 0.303 & 5.23 & 0.2798 & 0.2174 \\
\midrule
$10.0$ & $1$ & 23.15 & 0.3022 & 5.21 & 0.2782 & 0.2165 \\
$10.0$ & $2$ & 24.24 & 0.2067 & 5.21 & 0.2788 & 0.2166 \\
$10.0$ & $4$ & 26.55 & 0.3033 & 5.24 & 0.2801 & 0.2174 \\
\midrule
$12.5$ & $1$ & 23.75 & 0.3029 & 5.23 & 0.2791 & 0.2162  \\
$12.5$ & $2$ & 24.56 & 0.3032 & 5.24 & 0.2795 & 0.2168 \\
$12.5$ & $4$ & 26.64 & 0.3036 & 5.25 & 0.2804 & \textbf{0.2175} \\
\midrule
$15.0$ & $1$ & 24.37 & 0.3033 & 5.25 & 0.2798 & 0.2167 \\
$15.0$ & $2$ & 24.88 & 0.3036 & 5.25 & 0.28 & 0.2169 \\
$15.0$ & $4$ & 26.38 & \textbf{0.3038} & \textbf{5.26} & \textbf{0.2807} & \textbf{0.2175} \\
\bottomrule
\end{tabular}
\label{table:small-ms-coco-results-ablation-clg-128-steps}
\end{table}

\subsection{Ablation on time distribution $p(s,t)$ from~\Cref{sec:guidance_learning_algo}}
\label{sec_app:time_dist_ablations_main}

In~\Cref{table:small-ms-coco-results-ablation-sampling}, we present results with different hyperparameters of the time distribution $p(s,t)$ discussed in~\Cref{sec:guidance_learning_algo}. We see that overall the performance of the method is quite robust to these parameters -- the CLIP score does not seem to be affected much by these parameters.

\begin{table}[htbp]
\centering
\caption{Performance metrics for MMDiT-XS for different values of $\delta$ values in time distribution $p(s,t)$ and different CLIP reward scale $\eta$.}
\label{table:mmdit-metrics}
\small %
\begin{tabular}{cccccccc} 
\toprule
$\bm{\delta}$ & Reward scale $\eta$ & \textbf{FID} $\downarrow$ & \textbf{CLIP} $\uparrow$ & \textbf{Aesthetic} $\uparrow$  & \textbf{HPSv2} $\uparrow$ & \textbf{Pick Score} $\uparrow$ \\
\midrule
$0.01$ & 2.50 & 31.54 & 0.3049 & 5.3287 & 0.2830 & 0.2188 \\
$0.10$ & 2.50  & 31.15 & 0.3047 & 5.3145 & 0.2826 & 0.2188 \\
$0.20$ & 2.50  & 30.88 & 0.3049 & 5.3190 & 0.2828 & 0.2187 \\
$0.30$ & 2.50  & 31.16 & 0.3049 & 5.3207 & 0.2829 & 0.2189 \\ 
\addlinespace
$0.01$ & 0.25  & 31.24 & 0.3048 & 5.3179 & 0.2827 & 0.2189 \\
$0.10$ & 0.25   & 31.31 & 0.3048 & 5.3236 & 0.2829 & 0.2186 \\
$0.20$ & 0.25   & 30.36 & 0.3049 & 5.3118 & 0.2826 & 0.2185 \\
$0.30$ & 0.25   & 31.17 & 0.3049 & 5.3202 & 0.2828 & 0.2189 \\ 
\bottomrule
\end{tabular}
\label{table:small-ms-coco-results-ablation-sampling}
\end{table}

\subsection{Impact of number of sampling steps}
\label{sec_app:number_of_sampling_steps}

In~\Cref{tab:mmdit_sampling_steps_study}, we show the impact of number of sampling steps on performance of our method. We see that overall, doing more sampling steps leads to better perceptual metrics and to a better prompt alignment.

\begin{table}[htbp]
\centering
\caption{Impact of number of sampling steps on the performance of our GAN-based guidance scheduler. We use MMDiT-S backbone.}
\begin{tabular}{cccccc}
\toprule
\textbf{Sampling steps} & \textbf{FID $\downarrow$} & \textbf{CLIP $\uparrow$} & \textbf{Aesthetic $\uparrow$} & \textbf{HPSv2 $\uparrow$} & \textbf{PickScore $\uparrow$} \\
\midrule
$128$ & 31.70 & \textbf{0.3068} & \textbf{5.34} & \textbf{0.28564} & \textbf{0.2214} \\
$64$  & 30.55 & \textbf{0.3068} & 5.32 & 0.2853 & 0.2212 \\
$32$  & \textbf{28.45} & 0.3062 & 5.31 & 0.2835 &  0.2196 \\
\bottomrule
\end{tabular}
\label{tab:mmdit_sampling_steps_study}
\end{table}

\subsection{Impact of generator architecture}
\label{sec_app:ablation_gen_arch}

In~\Cref{tab:mmdit_xs_raw_input_vs_stats}, we present results of our method where we vary the architecture of our generator. See~\Cref{app_sec:experimental_details} for more details on the generator architecture. We see, that light-weight MLP generator which receives the statistics of $x_s$, leads to overall better performance compared to a more complex Convolutional neural net which receives raw latents $x_s$.

\begin{table}[htbp]
\centering
\caption{Impact of generator architecture and choice of inputs (\ie raw $x_s$ vs statistics) with MMDiT-XS backbone.}
\begin{tabular}{lcccccc}
\toprule
\textbf{Architecture} & $x_s$ input type & \textbf{FID $\downarrow$} & \textbf{CLIP $\uparrow$} & \textbf{Aesthetic $\uparrow$} & \textbf{HPSv2 $\uparrow$} & \textbf{PickScore $\uparrow$} \\
\midrule
Convolutional & $x_s$ latents & \textbf{26.38} & 0.3041 & 5.26 & 0.2807 & 0.2170 \\
MLP & $x_s$ statistics & 31.31 & \textbf{0.3048} & \textbf{5.32} & \textbf{0.2829} & \textbf{0.2186} \\
\bottomrule
\end{tabular}
\label{tab:mmdit_xs_raw_input_vs_stats}
\end{table}

\subsection{Impact of how the noise is sampled during training}
\label{sec_app:noise_impact}

In \Cref{alg:flowmatching-gan-dre}, the real target $x_t^{\text{real}}$ and the proposal base $x_s$ can be constructed using either coupled ($\epsilon = z$) or independent ($\epsilon \neq z$) Gaussian noise vectors. 
We evaluate this design choice in \Cref{tab:mmdit_xs_noise_ablation} using an MMDiT-XS backbone. 
The empirical metrics indicate that enforcing true independence between the target and proposal noise states provides minor improvements in downstream generation quality (e.g., lower FID and improved HPSv2). 
This confirms that the marginal consistency objective is robust to the underlying noise coupling strategy, as the network generalizes effectively without requiring correlated trajectory pairings during training.
\label{sec_app:noise_ablation}
\begin{table}[htbp]
\centering
\caption{Impact of how the noise sampled on the performance of our GAN-based guidance scheduler. We use MMDiT-XS backbone.}
\begin{tabular}{cccccc}
\toprule
\textbf{Independent noise} & \textbf{FID $\downarrow$} & \textbf{CLIP $\uparrow$} & \textbf{Aesthetic $\uparrow$} & \textbf{HPSv2 $\uparrow$} & \textbf{PickScore $\uparrow$} \\
\midrule
True & \textbf{31.31} & \textbf{0.3048} & \textbf{5.32} & \textbf{0.2829} & \textbf{0.2186} \\
False  & 31.41 & \textbf{0.3048} & \textbf{5.32} & 0.2827 & \textbf{0.2186} \\
\bottomrule
\end{tabular}
\label{tab:mmdit_xs_noise_ablation}
\end{table}

\subsection{Impact of how image-conditioning pairs are sampled}
\label{sec_app:how_image_pairs_sampled}
The ablation study in~\Cref{tab:mmdit_xs_correlated_vs_independent} demonstrates that our method is quite robust to the sampling strategy of image-conditioning pairs. While independent sampling is theoretically more aligned with the marginal consistency objective, it yields nearly identical performance to correlated sampling across most metrics, with only a marginal improvement observed in HPSv2 ($0.2927$ vs. $0.2829$).

\begin{table}[htbp]
\centering
\caption{Impact of choosing independent image-conditioning pairs on performance of our method with MMDiT-XS backbone.}
\begin{tabular}{cccccc}
\toprule
 \textbf{Independent} $\bm{(x_1,c)}$ & \textbf{FID $\downarrow$} & \textbf{CLIP $\uparrow$} & \textbf{Aesthetic $\uparrow$} & \textbf{HPSv2 $\uparrow$} & \textbf{PickScore $\uparrow$} \\
\midrule
\checkmark & 31.41 & \textbf{0.3048} & \textbf{5.32} & 0.2927 & \textbf{0.2186} \\
\xmark & \textbf{31.31} & \textbf{0.3048} & \textbf{5.32} & \textbf{0.2829} & \textbf{0.2186} \\
\bottomrule
\end{tabular}
\label{tab:mmdit_xs_correlated_vs_independent}
\end{table}

\subsection{Impact of CLIP reward loss}
We perform an ablation of our method without CLIP reward loss~\Cref{eq:clip_reward_loss}. We notice that the metrics such as CLIP score and human preference metrics drop marginally in the absence of this loss for MMDiT-XS model. We summarize these results in \Cref{tab:mmdit_xs_clip_loss_ablation}.

\begin{table}[h]
\centering
\caption{Impact of choosing independent image-conditioning pairs on performance of our method with MMDiT-XS backbone.}
\begin{tabular}{cccccc}
\toprule
\textbf{CLIP reward} & \textbf{FID $\downarrow$} & \textbf{CLIP $\uparrow$} & \textbf{Aesthetic $\uparrow$} & \textbf{HPSv2 $\uparrow$} & \textbf{PickScore $\uparrow$} \\
\midrule
\checkmark & 31.41 & \textbf{0.3048} & \textbf{5.32} & \textbf{0.2927} & \textbf{0.2186} \\
\xmark & \textbf{31.11} & 0.3047 & 5.31 & 0.2825 & 0.2184 \\
\bottomrule
\end{tabular}
\label{tab:mmdit_xs_clip_loss_ablation}
\end{table}

\subsection{Alternative choices for the distribution $p(s,t)$}
\label{sec_app:other_time_dist}

The distribution $p(s,t)$ from Section~\ref{sec:guidance_learning_algo} samples $t$ uniformly and then draws the step size $\Delta s$ uniformly conditional on $t$. The set of valid time pairs forms a triangular (trapezoid) region
\begin{equation}
    \mathcal{T} = \bigl\{(s, t) \;:\; \zeta + \delta \le t \le 1 - \zeta,\;\; \zeta \le s \le t - \delta \bigr\}.
    \label{eq:triangular_support}
  \end{equation}
However, $p(s,t)$ from Section~\ref{sec:guidance_learning_algo} does not lead to a uniform distribution over this region.

    \paragraph{Uniform distribution over $\mathcal{T}$.}
  We now discuss a choice where we sample uniformly over $\mathcal{T}$. In order to achieve this, we: (1)~draw $u \sim \mathcal{U}[0,1]$; (2)~compute $t = \zeta + \delta + (1 - 2\zeta - \delta)\,\sqrt{u}$; and (3)~sample $s \sim \mathcal{U}[\zeta,\, t - \delta]$.

  Now, we will write a quick derivation for that. A uniform joint density over $\mathcal{T}$ requires $p(s,t) = 1/|\mathcal{T}|$ for all $(s,t) \in \mathcal{T}$.
  Factoring the joint as $p(s,t) = p(t)\,p(s \mid t)$ and choosing $s \mid t \sim \mathcal{U}[\zeta,\, t - \delta]$, i.e.\ $p(s \mid t) = 1/w(t)$ where $w(t) = t - \delta - \zeta$ is the width of the valid strip at~$t$, we obtain
  \begin{equation}
    p(s,t) = p(t) \cdot \frac{1}{w(t)}.
  \end{equation}
  Setting this equal to $1/|\mathcal{T}|$ and solving for $p(t)$ gives
  \begin{equation}
    p(t) = \frac{w(t)}{|\mathcal{T}|} = \frac{t - \delta - \zeta}{|\mathcal{T}|},
    \label{eq:marginal_t}
  \end{equation}
  confirming that $t$ must be sampled with probability proportional to the strip width $w(t)$.
  The area of $\mathcal{T}$ is
  \begin{equation}
    |\mathcal{T}| = \int_{\zeta + \delta}^{1 - \zeta} w(t)\,\mathrm{d}t
    = \int_{\zeta + \delta}^{1 - \zeta} (t - \delta - \zeta)\,\mathrm{d}t
    = \frac{(1 - 2\zeta - \delta)^2}{2}.
  \end{equation}
  The CDF of~$t$ under~\eqref{eq:marginal_t} is
  \begin{equation}
    F(t) = \frac{(t - \zeta - \delta)^2}{(1 - 2\zeta - \delta)^2},
  \end{equation}
  which inverts to
  \begin{equation}
    t = \zeta + \delta + (1 - 2\zeta - \delta)\,\sqrt{u}, \qquad u \sim \mathcal{U}[0,1].
    \label{eq:inverse_cdf}
  \end{equation}
  Given~$t$, the source time $s \mid t \sim \mathcal{U}[\zeta,\, t - \delta]$.
  Substituting back, the joint density is
  $p(s,t) = p(t) / w(t) = 1/|\mathcal{T}|$,
  which is constant over $\mathcal{T}$.

  \paragraph{Beta-trapezoid distribution over $\mathcal{T}$.}
  The uniform-joint sampler above can be generalized by replacing the uniform quantile $u \sim \mathcal{U}[0,1]$ with a Beta-distributed one,
  $u \sim \mathrm{Beta}(\alpha, \beta)$,
  and feeding it through the same inverse CDF~\eqref{eq:inverse_cdf}.
  When $\alpha = \beta = 1$, the Beta distribution reduces to $\mathcal{U}[0,1]$, exactly recovering the uniform-joint distribution.
  Since the inverse CDF $t = \zeta + \delta + (1 - 2\zeta - \delta)\,\sqrt{u}$ is monotonically increasing in $u$, biasing $u$ toward larger values translates directly into biasing $t$ toward larger values.
  Setting $\alpha > \beta$ increases the mean of the Beta distribution (recall $\mathbb{E}[u] = \alpha / (\alpha + \beta)$), which shifts mass toward larger $t$ (closer to data), while $\beta > \alpha$ shifts mass toward smaller $t$ (closer to noise).

  \paragraph{Results.} We present the results in~\Cref{tab:beta_trapezoid_study}. The last 3 rows represent the distribution $p(s,t)$ from Section~\ref{sec:guidance_learning_algo}. We see that overall $\beta < \alpha$, leads to worse results, while $\alpha\geq\beta$ performs better. Overall, $\alpha=1, \beta=2$ leads to the best results, which are comparable to the ones obtained with  the distribution $p(s,t)$ from Section~\ref{sec:guidance_learning_algo}.

  \begin{table}[htbp!]
  \centering
  \caption{\textbf{Beta-trapezoid time distribution ablation on MMDiT-XS.} We vary the Beta distribution parameters $(\alpha, \beta)$ and minimum step size $\delta$. When $\alpha = \beta = 1$, the distribution reduces to uniform sampling over the triangular support~$\mathcal{T}$. The last 3 rows correspond to the distribution $p(s,t)$ from Section~\ref{sec:guidance_learning_algo}.}
  \begin{tabular}{cccccccc}
  \toprule
  $\bm{\alpha}$ & $\bm{\beta}$ & $\bm{\delta}$ & \textbf{FID} $\downarrow$ & \textbf{CLIP} $\uparrow$ & \textbf{Aesthetic} $\uparrow$ & \textbf{HPSv2} $\uparrow$  & \textbf{PickScore $\uparrow$} \\
  \midrule
  \multirow{3}{*}{1} & \multirow{3}{*}{1}
    & 0.01 & 30.85 & 0.3048 & 5.32 & 0.2827 & 0.21847 \\
      & & 0.10  & 30.79 & 0.3038 & 5.31 & 0.2819 & 0.21786 \\
      & & 0.20  & 31.52 & 0.3048 & 5.325 & 0.2828 & 0.21861\\
      \midrule
      \multirow{3}{*}{1} & \multirow{3}{*}{2}
        & 0.01 & 31.87 & 0.3046 & 5.32 & 0.2826 & 0.21869 \\
      & & 0.10  & 32.00 & 0.3048 & 5.33 & 0.2829 & 0.21882\\
      & & 0.20  & 31.54 & 0.3048 & 5.32 & 0.2827 & 0.21851 \\
      \midrule
      \multirow{3}{*}{2} & \multirow{3}{*}{1}
        & 0.01 & 30.58 & 0.3033 & 5.31 & 0.2824 & 0.21831 \\
      & & 0.10  & 28.71 & 0.3013 & 5.26 & 0.2795 & 0.21649 \\
      & & 0.20  & 30.57 & 0.3033 & 5.30 & 0.2814 & 0.21762 \\
      \midrule
      \multirow{3}{*}{-} & \multirow{3}{*}{-}
        & $0.01$ & 31.20 & 0.30476 & 5.32 & 0.2827 & 0.21888 \\
        & & $0.10$  & 31.31 & 0.30484 & 5.32 & 0.2829 & 0.21862 \\
        & & $0.20$  & 30.30 & 0.30486 & 5.31 & 0.2826 & 0.21850 \\
  \bottomrule
  \end{tabular}
  \label{tab:beta_trapezoid_study}
  \end{table}

\section{Experimental Details}
\label{app_sec:experimental_details}
We specify architectural details of discriminator and guidance network below. in addition, we state our optimizer hyperparameter settings in \Cref{table:hyperparameter-details}.

\paragraph{Discriminator architecture.}
The discriminator $d_\phi(x_t, s, t, c)$ is a convolutional network inspired by DCGAN architecture \citep{radford2015unsupervised}. 
It follows a modular downsampling structure consisting of five primary modules. Each module utilizes a Conv-Residual-AdaGN-Leaky-ReLU block. 
For convolutional layers, we utilize $4 \times 4$ kernels with a stride of 2 and explicit padding of 1 for the first three stages to progressively reduce spatial resolution while doubling feature depth. The final two stages utilize $2 \times 2$ kernels to reach a $1 \times 1$ spatial bottleneck.
Each block incorporates a residual connection with a $1 \times 1$ projection convolution to match feature dimensions.
We also apply spectral normalization \citep{miyato2018spectral} to all convolutional layers to enforce Lipschitz continuity and prevent GAN instability.
The discriminator starts with a base feature width of 64, expanding to 1024 in the final layer. The Leaky ReLU uses slope of 0.2.

The network also accepts two time steps, the source time $s$ and target time $t$. 
Each temporal scalar is projected into a 64-dimensional space via separate MLP-based time encoders.
These embeddings are concatenated and processed through a two-layer MLP with SiLU activation to create a joint temporal embedding $\mathbf{e}_{\text{joint}}$.
The CLIP text embeddings $c_{\text{CLIP}}$ are projected via a dedicated MLP to match the hidden dimension of the temporal features.
These three embeddings are concatenated and passed through a two layer MLP with SiLU activations, and then injected into each convolutional block using Adaptive Group Normalization (AdaGN). 

We use the same discriminator architecture for both MMDiT-XS and MMDiT-S. Overall, this discriminator has 5M trainable parameters. We also use this same architecture for the generator architecture for the results in \Cref{tab:mmdit_xs_raw_input_vs_stats}.

\paragraph{Guidance network architecture.} The guidance network $\boldsymbol{\omega}(x_s,s,t,c;\psi)$ is a MLP that predicts guidance scale. 
We utilize Fourier Embeddings to project the source time $s$, target time $t$, and the temporal gap $(s-t)$ into a 256-dimensional space. 
We use standard sinusoidal frequencies with a $1/10000$ scaling factor. 
These embeddings are concatenated and passed through a linear layer with Layer Normalization and SiLU activation.
We also input the CLIP text embeddings after projecting them into a 256-dimensional hidden space via a Dense-LayerNorm-SiLU block.
We also condition the network on a 5-dimensional vector of coordinate-invariant geometric statistics as described in \cref{sec:guidance_learning_algo} of the main paper.
These statistics are projected to a 128-dimensional feature vector.
The fused features are passed into a central backbone consisting of a projection layer followed by two Residual Blocks. Each residual block follows the structure:
\begin{align}
\text{Output} = \text{SiLU}(x + \text{LayerNorm}(\text{Dense}(\text{SiLU}(\text{LayerNorm}(\text{Dense}(x))))))
\end{align}
The final layer is a linear projection to a single scalar, followed by a softplus activation to ensure non-negative guidance scale.
We utilize a zero-initialization strategy for the final weight matrix. 
The bias of the final layer is initialized to $\text{softplus}^{-1}(\alpha)$, where $\alpha$ is the starting guidance multiplier (e.g., 1.0). This ensures that at the start of training, the model defaults to a standard constant schedule before learning to optimize the trajectory.

\begin{table}[h]
\centering
\caption{\textbf{Training and Sampling Details.} Hyperparameter settings for MMDiT-XS, MMDiT-S.
}
\begin{tabular}{lccc}
\toprule
\textbf{Config} & \textbf{MMDiT-XS} & \textbf{MMDiT-S} \\
\midrule
\textit{Discriminator Training} \\
\midrule
Batch size & 128 & 128 \\
Optimizer & Adam & Adam \\
Learning rate & 1e-5 & 1e-5 \\
$(\beta_1, \beta_2)$ & (0.5, 0.99) & (0.5, 0.99)\\
Warmup steps & $5000$ & 5000\\
Training steps & $60000$ & $60000$ \\
$R_1$ penalty $\gamma$ & 0.1 & 0.1 \\
$R_1$ penalty interval & 5 & 4 \\
\midrule
\textit{Generator Training} \\
\midrule
Batch size & 128 & 128 \\
Optimizer & Adam & Adam \\
Learning rate & 2e-5 & 2e-5 \\
$(\beta_1, \beta_2)$ & (0.9, 0.999) & (0.9, 0.999) \\
Warmup steps & $5000$ & $5000$ \\
Training steps & $60000$ & $60000$ \\
Regularization weight $a$ & 0.1 & 1e-5 \\
Clip reward weight $\eta$ & 0.25 & 0.25 \\
\midrule
\textit{Sampling} \\
\midrule
Sampler type & Euler & Euler \\
Sampling steps & 128 & 128 \\
\bottomrule
\end{tabular}
\label{table:hyperparameter-details}
\end{table}

\paragraph{Training details.}
The training details are specified in~\Cref{table:hyperparameter-details}. All models are trained for $60000$ iterations. During training, we track CLIP score on a subset of $3000$ images from MS-COCO $512\times512$. We use this metric to select the best checkpoint for every studied method.

\paragraph{Baselines training details.} For MMD+SC and MMD+MC baselines, unless specified otherwise, we follow the same training protocol as above and use the guidance network $\boldsymbol{\omega}(x_s,s,t,c)$. The only difference is that we do not employ discriminator and we do not use regularization~\eqref{eq:reg_term}. On top of that, we consider the same experimental setup where we do not include $x_s$ into the guidance network $\boldsymbol{\omega}(s,t,c)$. Finally, we also include results with $\boldsymbol{\omega}(s,t,c)^\dagger$, where we follow the setup from \citet{galashov2025learn}.

\section{Image samples}
\label{sec_app:image_samples}

\begin{figure}[h]
\centering
\begin{subfigure}{.95\textwidth}
    \includegraphics[width=0.19\textwidth]{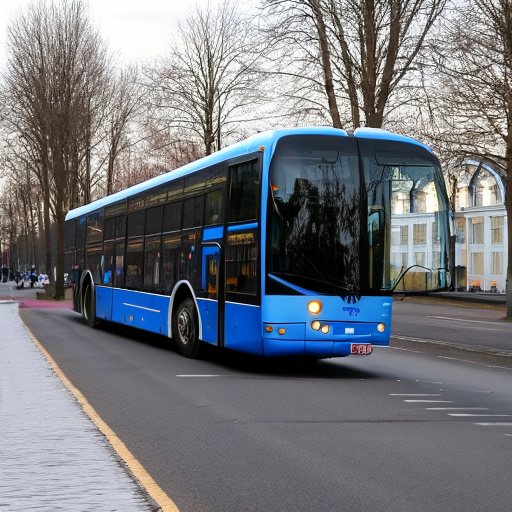}
    \includegraphics[width=0.19\textwidth]{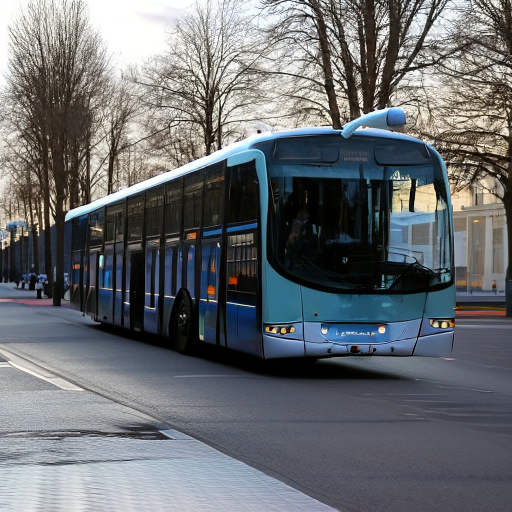}
    \includegraphics[width=0.19\textwidth]{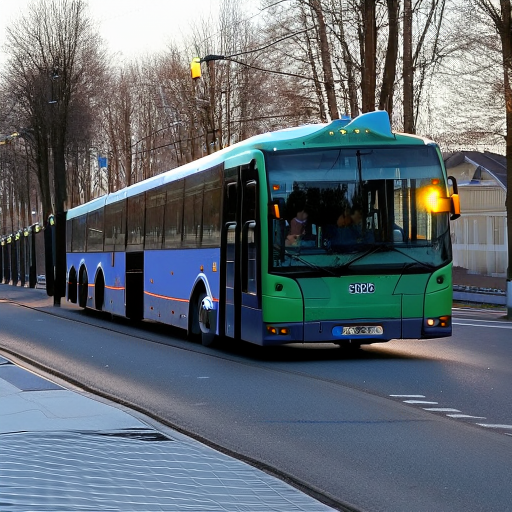}
    \includegraphics[width=0.19\textwidth]{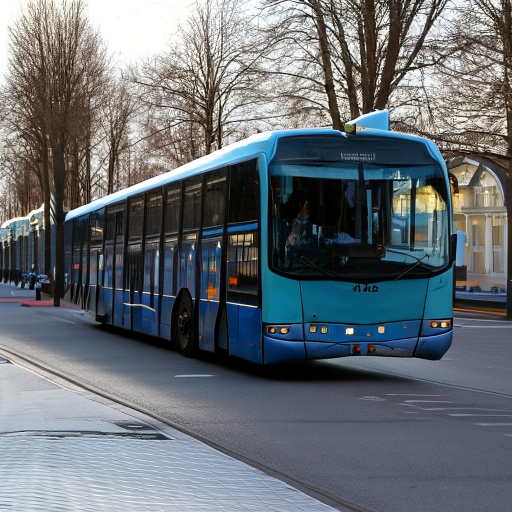}
    \includegraphics[width=0.19\textwidth]{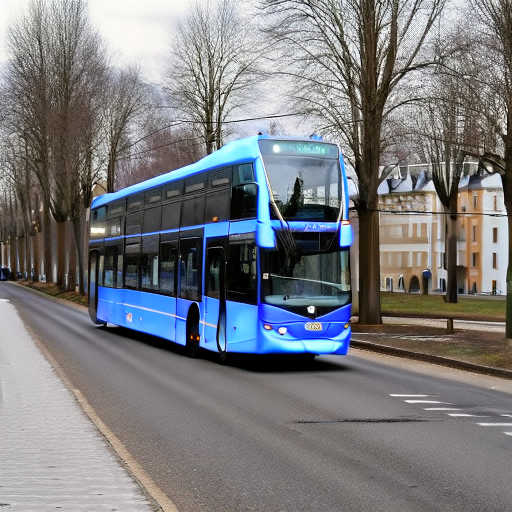}
    \caption{\textit{A city bus is riding down the empty street.}}
\end{subfigure}
\hfill
\begin{subfigure}{.95\textwidth}
    \includegraphics[width=0.19\textwidth]{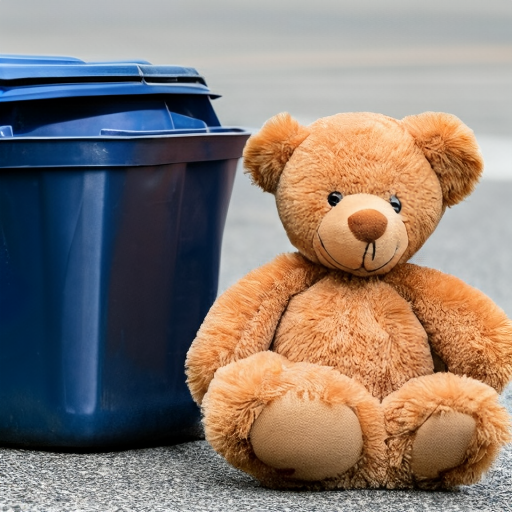}
    \includegraphics[width=0.19\textwidth]{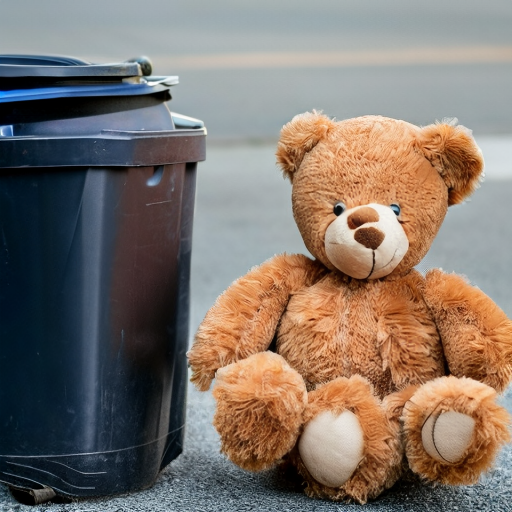}
    \includegraphics[width=0.19\textwidth]{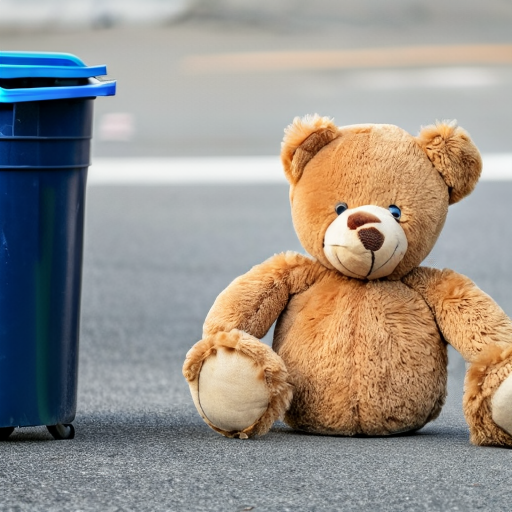}
    \includegraphics[width=0.19\textwidth]{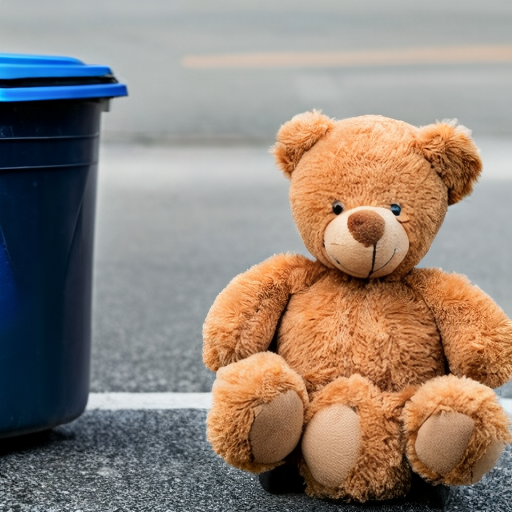}
    \includegraphics[width=0.19\textwidth]{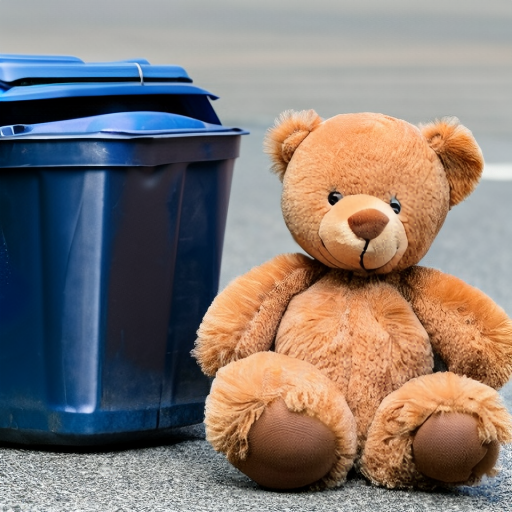}
    \caption{\textit{Stuffed teddy bear sitting next to garbage can on the side of the road.}}
\end{subfigure}
\hfill
\begin{subfigure}{.95\textwidth}
    \includegraphics[width=0.19\textwidth]{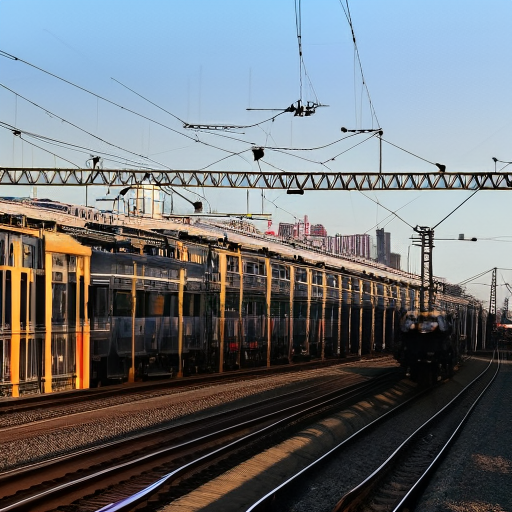}
    \includegraphics[width=0.19\textwidth]{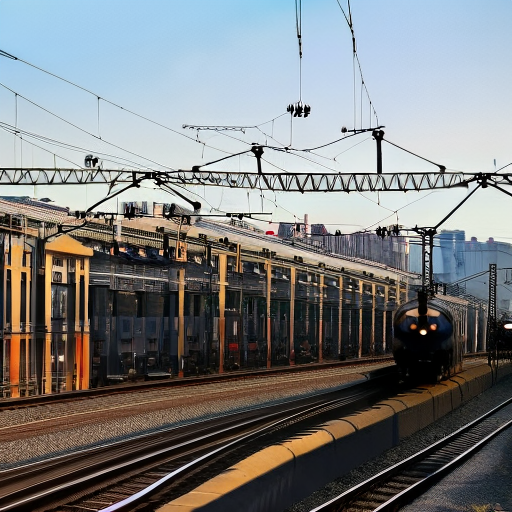}
    \includegraphics[width=0.19\textwidth]{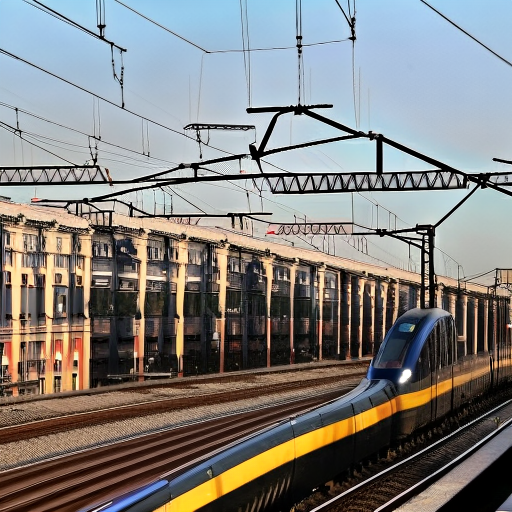}
    \includegraphics[width=0.19\textwidth]{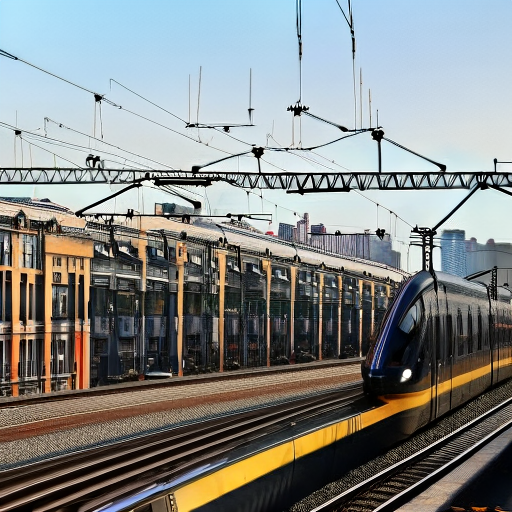}
    \includegraphics[width=0.19\textwidth]{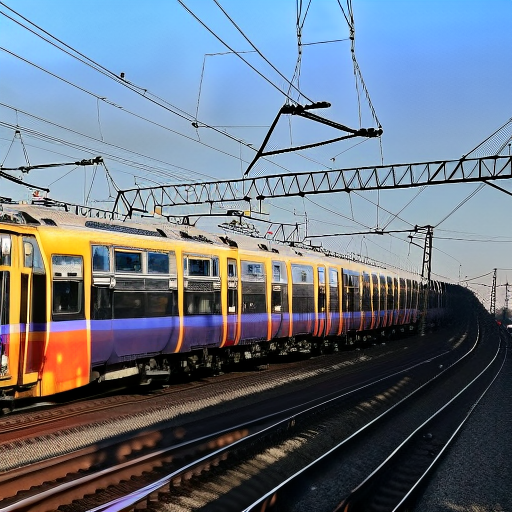}
    \caption{\textit{The train moves thru this part of the city.}}
\end{subfigure}
\hfill
\begin{subfigure}{.95\textwidth}
    \includegraphics[width=0.19\textwidth]{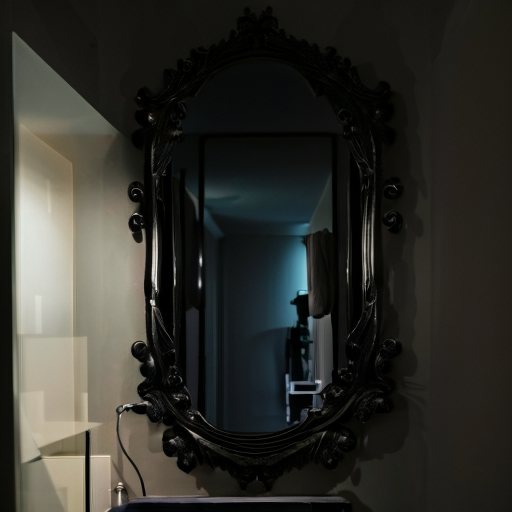}
    \includegraphics[width=0.19\textwidth]{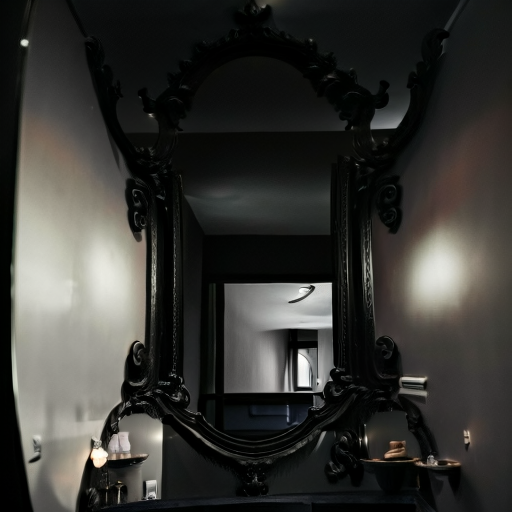}
    \includegraphics[width=0.19\textwidth]{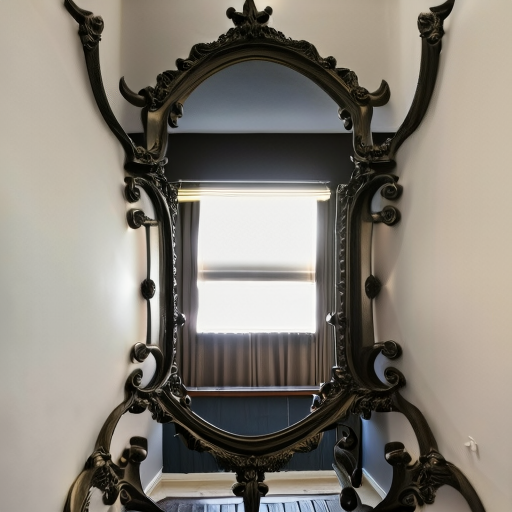}
    \includegraphics[width=0.19\textwidth]{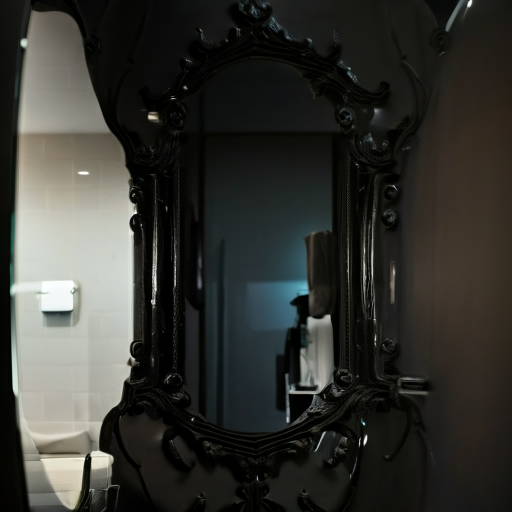}
    \includegraphics[width=0.19\textwidth]{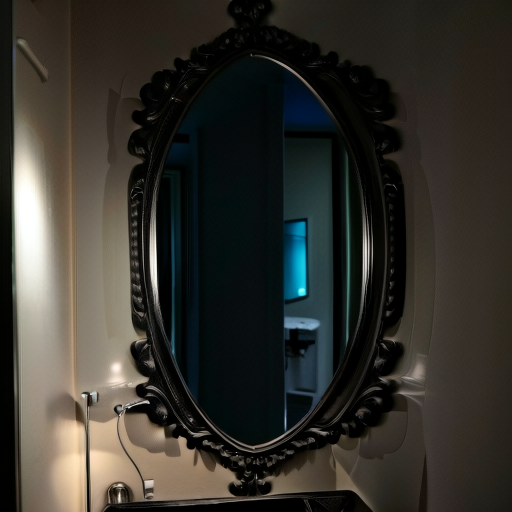}
    \caption{\textit{The dark mirror shows a bathroom in its reflection.}}
\end{subfigure}
\hfill
\begin{subfigure}{.95\textwidth}
    \includegraphics[width=0.19\textwidth]{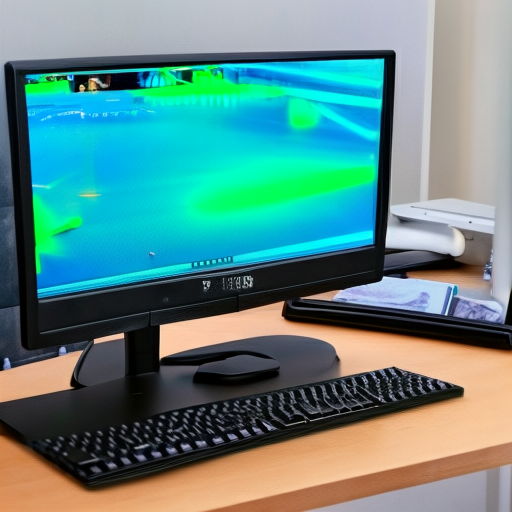}
    \includegraphics[width=0.19\textwidth]{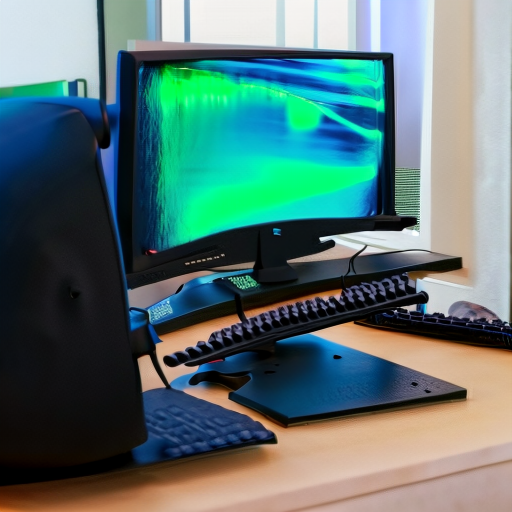}
    \includegraphics[width=0.19\textwidth]{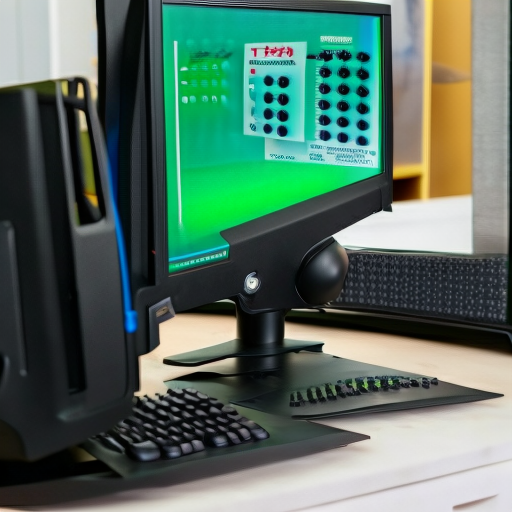}
    \includegraphics[width=0.19\textwidth]{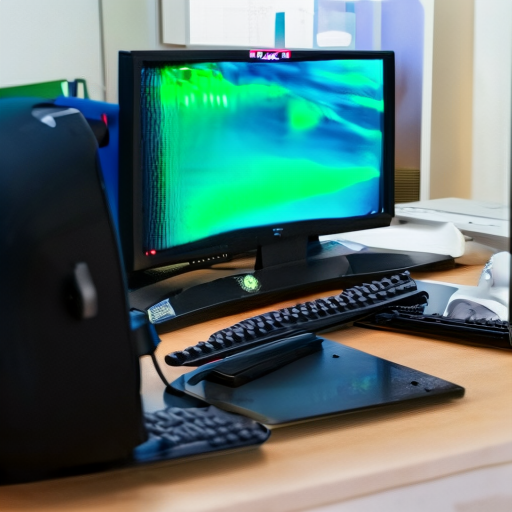}
    \includegraphics[width=0.19\textwidth]{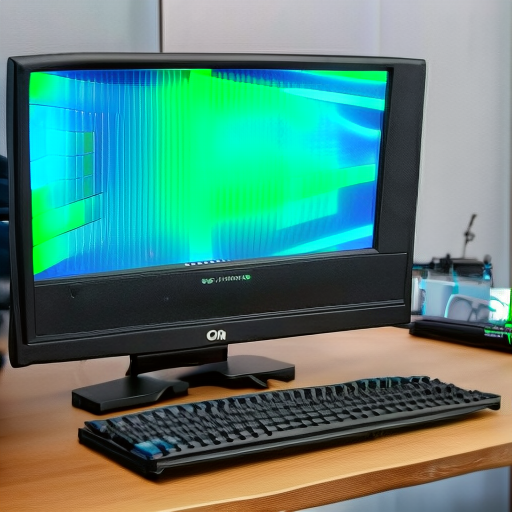}
    \caption{\textit{A computer monitor sitting on top of a computer desk.}}
\end{subfigure}
\hfill
\begin{subfigure}{.95\textwidth}
    \includegraphics[width=0.19\textwidth]{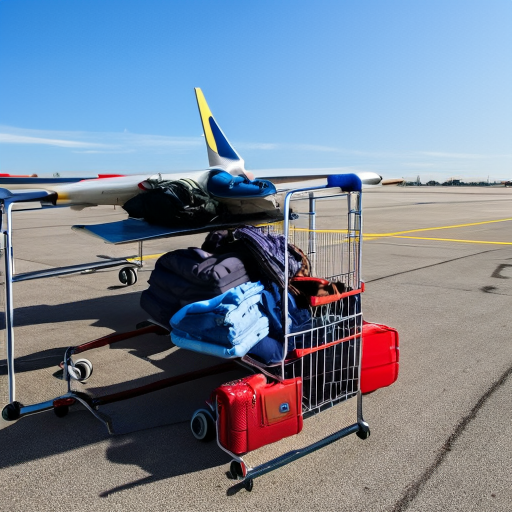}
    \includegraphics[width=0.19\textwidth]{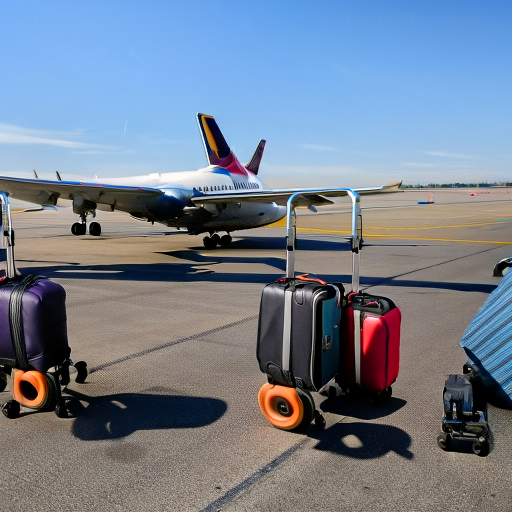}
    \includegraphics[width=0.19\textwidth]{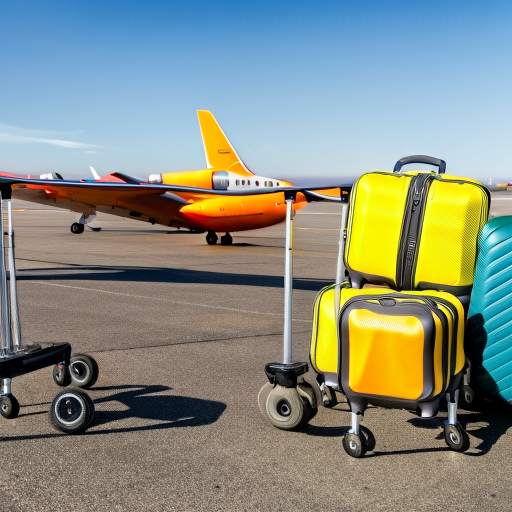}
    \includegraphics[width=0.19\textwidth]{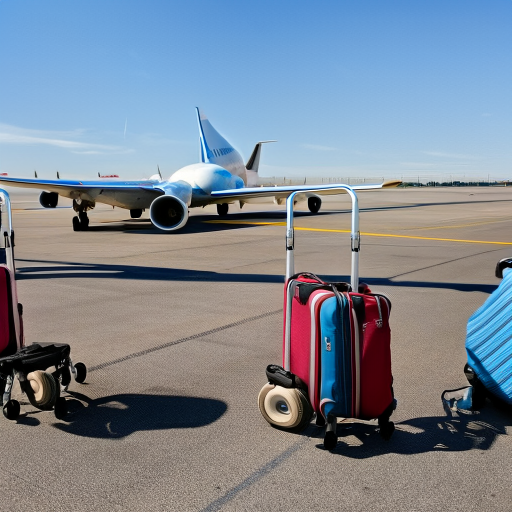}
    \includegraphics[width=0.19\textwidth]{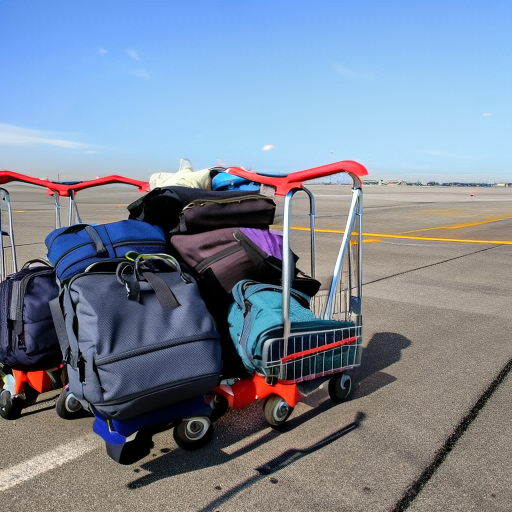}
    \caption{\textit{Luggage sitting on and around airline luggage cards on the tarmac.}}
\end{subfigure}
\hfill
\begin{subfigure}{.95\textwidth}
    \includegraphics[width=0.19\textwidth]{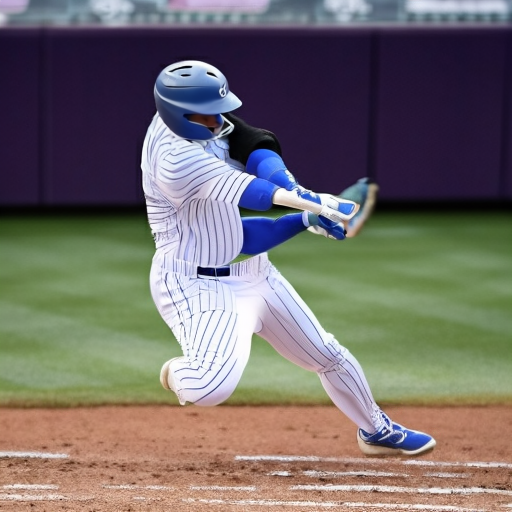}
    \includegraphics[width=0.19\textwidth]{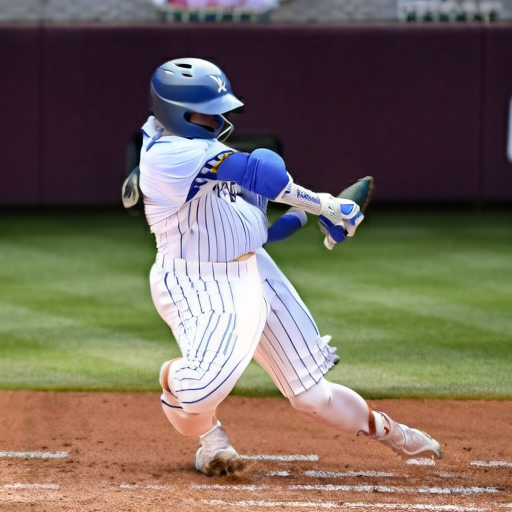}
    \includegraphics[width=0.19\textwidth]{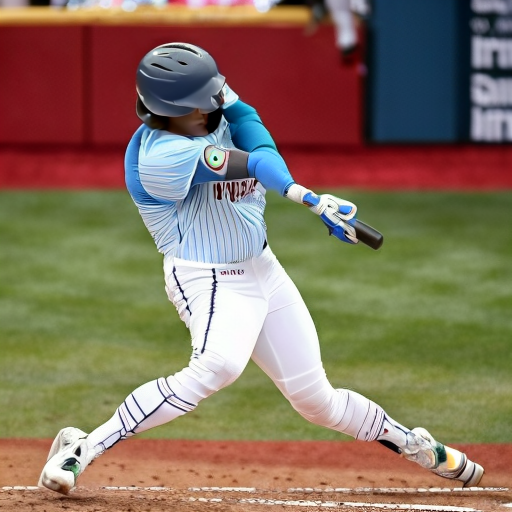}
    \includegraphics[width=0.19\textwidth]{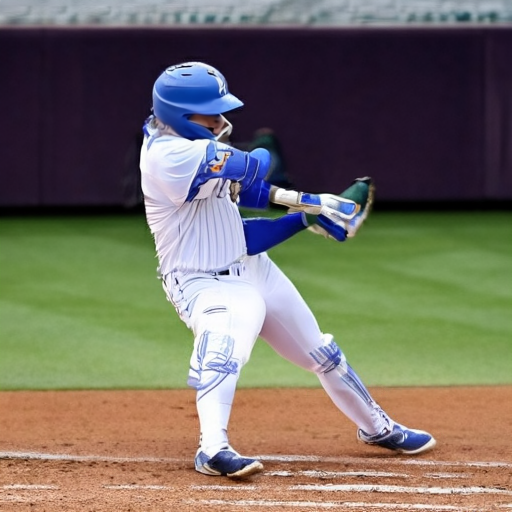}
    \includegraphics[width=0.19\textwidth]{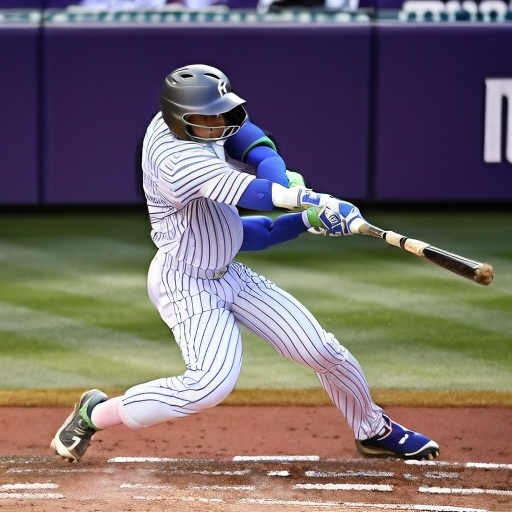}
    \caption{\textit{A batter swings at the ball during a baseball game.}}
\end{subfigure}
\caption{\textbf{T2I Results on MS-COCO with MMDiT-XS}. (left-to-right) We provide results of images generated from the given text prompt with CFG $\omega = 7.5$, CLG, LIG, MMD and our method.}
\label{fig:small-ms-coco-visualization-additional}
\end{figure}

\begin{figure}[h]
\centering
\begin{subfigure}{.95\textwidth}
    \includegraphics[width=0.23\textwidth]{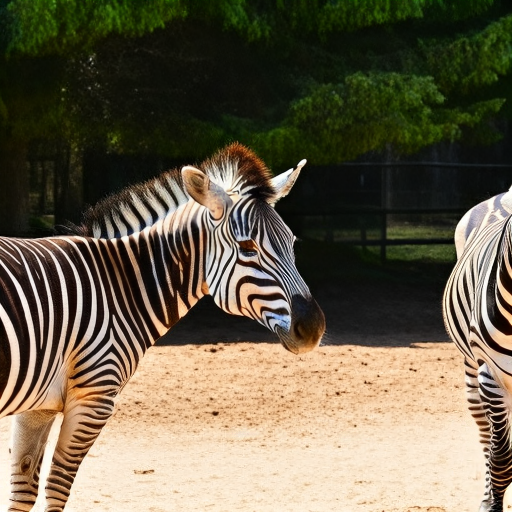}
    \includegraphics[width=0.23\textwidth]{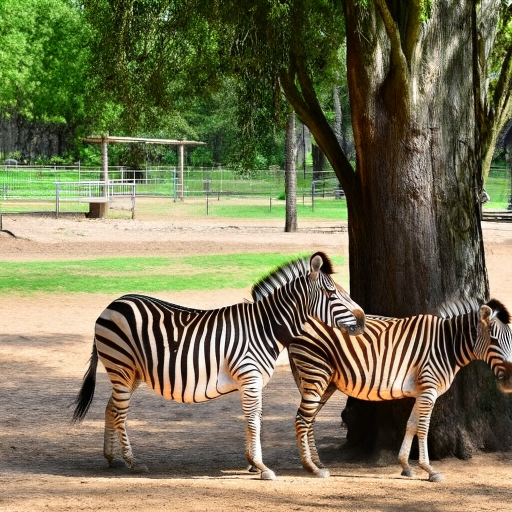}
    \includegraphics[width=0.23\textwidth]{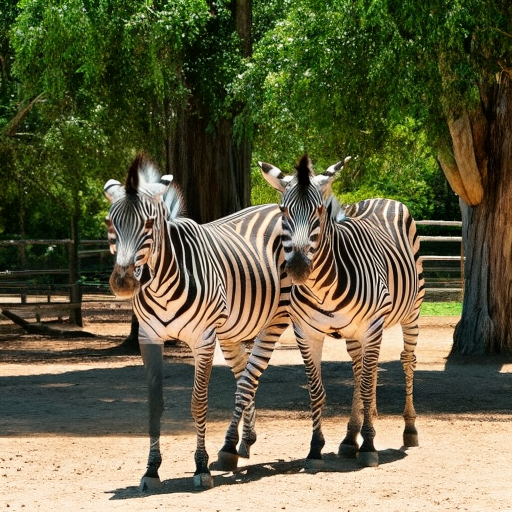}
    \includegraphics[width=0.23\textwidth]{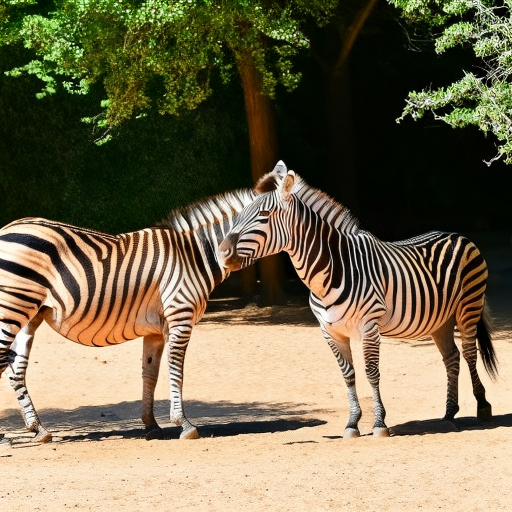}
    \caption{\textit{Two zebras in the zoo by some trees.}}
\end{subfigure}
\hfill
\begin{subfigure}{.95\textwidth}
    \includegraphics[width=0.23\textwidth]{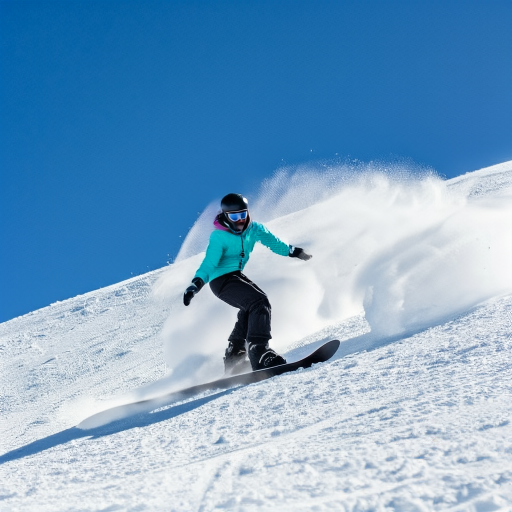}
    \includegraphics[width=0.23\textwidth]{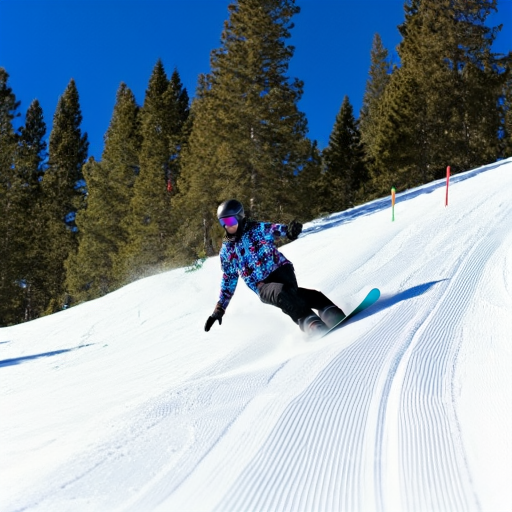}
    \includegraphics[width=0.23\textwidth]{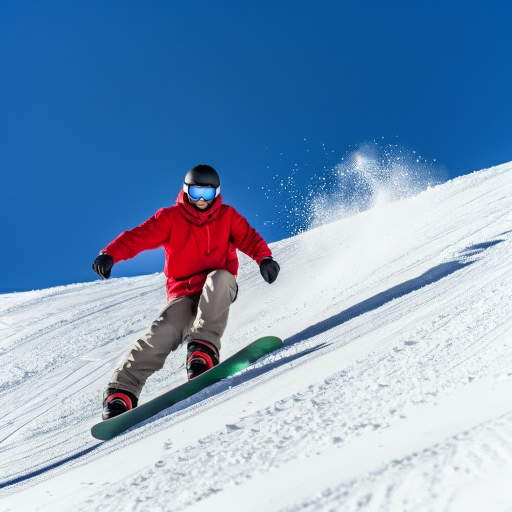}
    \includegraphics[width=0.23\textwidth]{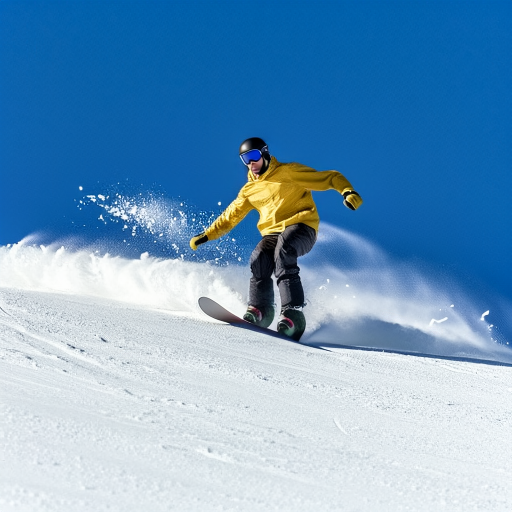}
    \caption{\textit{A man riding a snowboard down the side of a snow covered ski slope.}}
\end{subfigure}
\hfill
\begin{subfigure}{.95\textwidth}
    \includegraphics[width=0.23\textwidth]{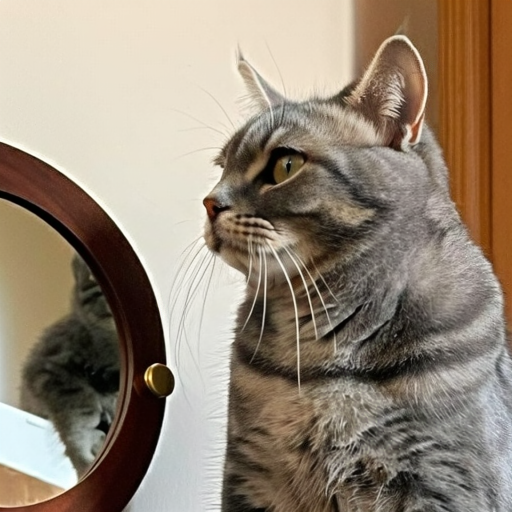}
    \includegraphics[width=0.23\textwidth]{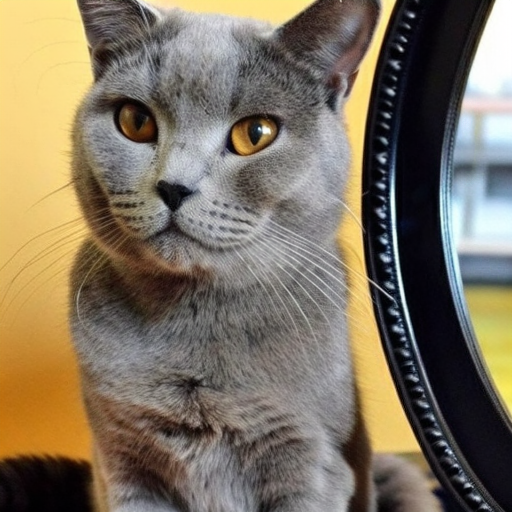}
    \includegraphics[width=0.23\textwidth]{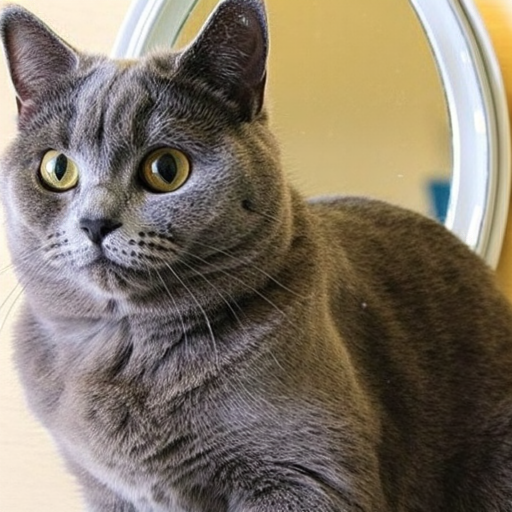}
    \includegraphics[width=0.23\textwidth]{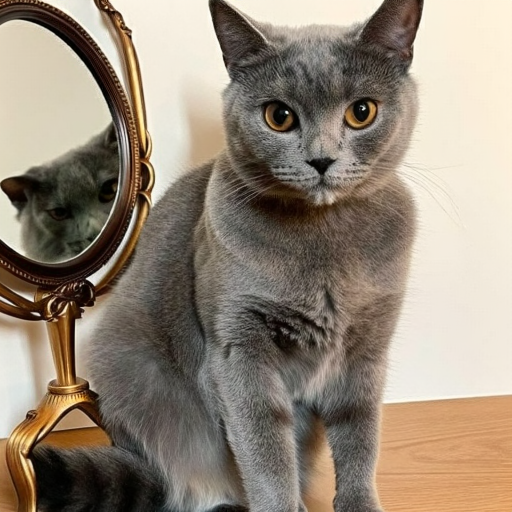}
    \caption{\textit{A grey cat sitting by a round mirror.}}
\end{subfigure}
\hfill
\begin{subfigure}{.95\textwidth}
    \includegraphics[width=0.23\textwidth]{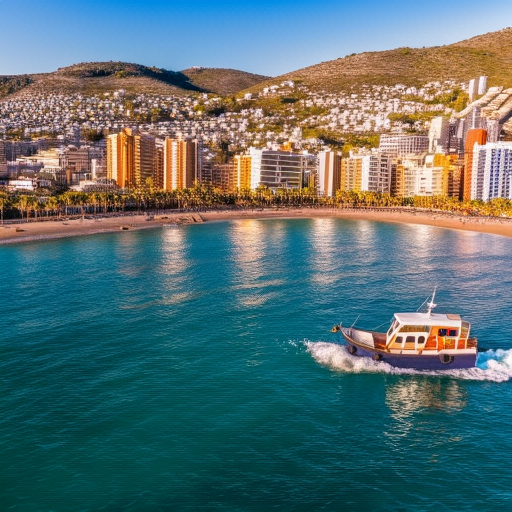}
    \includegraphics[width=0.23\textwidth]{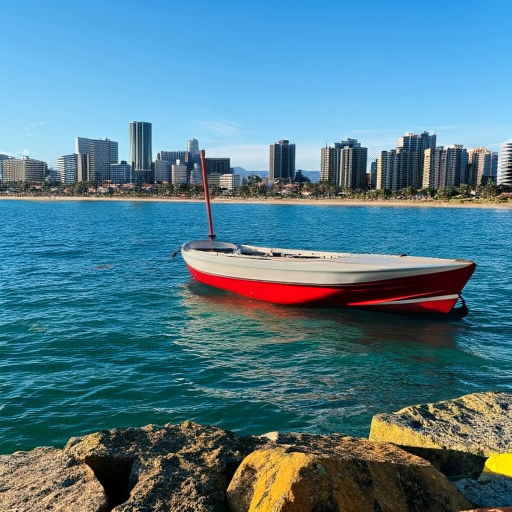}
    \includegraphics[width=0.23\textwidth]{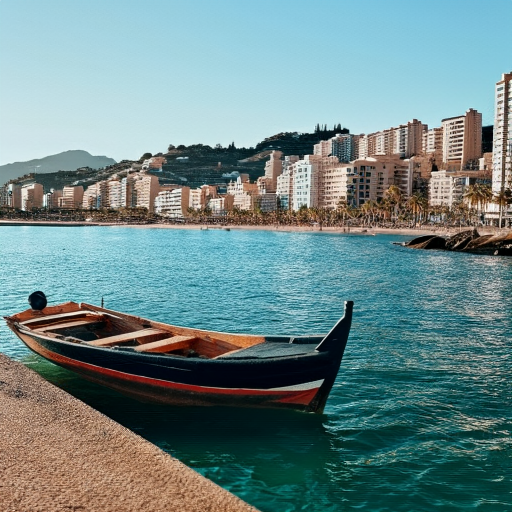}
    \includegraphics[width=0.23\textwidth]{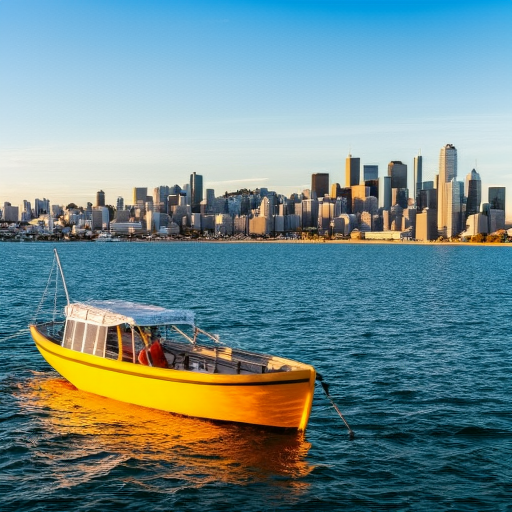}
    \caption{\textit{A small board sits on the water next to a city.}}
\end{subfigure}
\hfill
\begin{subfigure}{.95\textwidth}
    \includegraphics[width=0.23\textwidth]{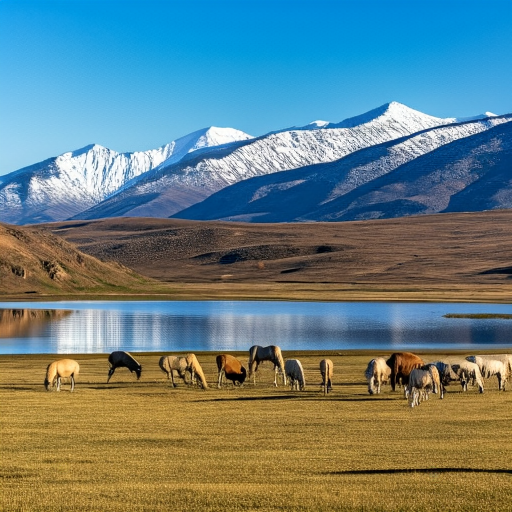}
    \includegraphics[width=0.23\textwidth]{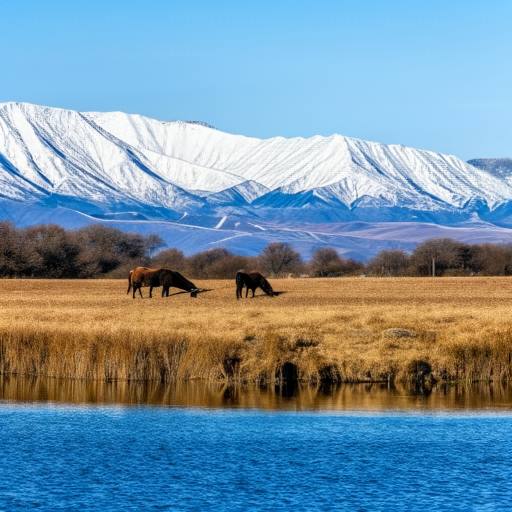}
    \includegraphics[width=0.23\textwidth]{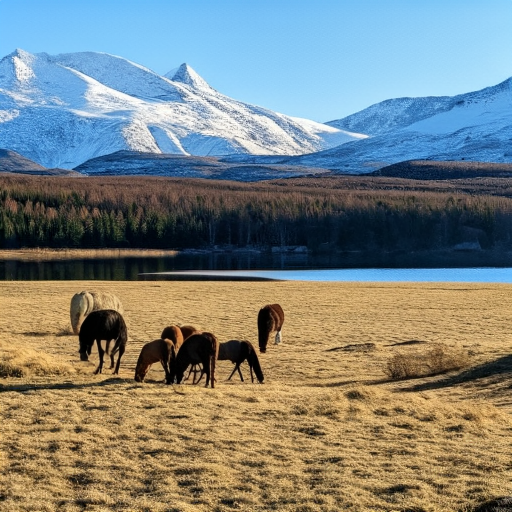}
    \includegraphics[width=0.23\textwidth]{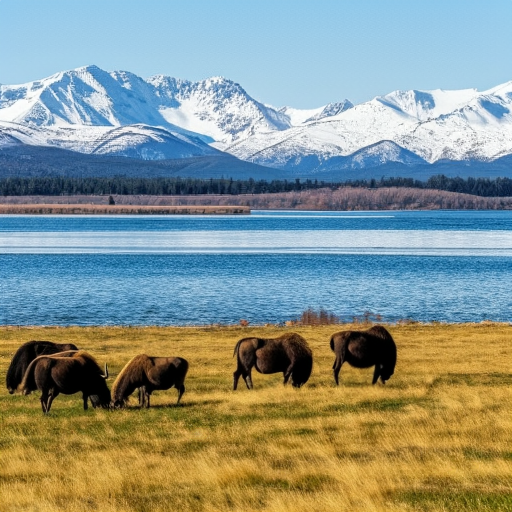}
    \caption{\textit{Wild animals graze in field in front of a lake and snow covered mountains.}}
\end{subfigure}
\hfill
\begin{subfigure}{.95\textwidth}
    \includegraphics[width=0.23\textwidth]{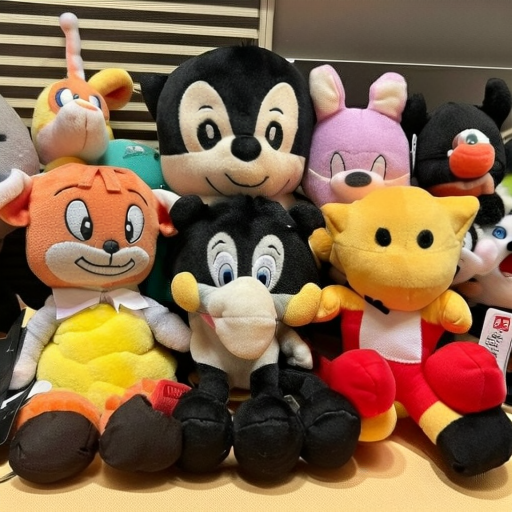}
    \includegraphics[width=0.23\textwidth]{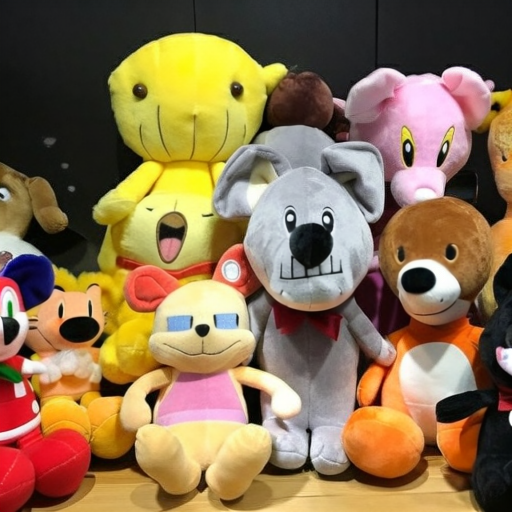}
    \includegraphics[width=0.23\textwidth]{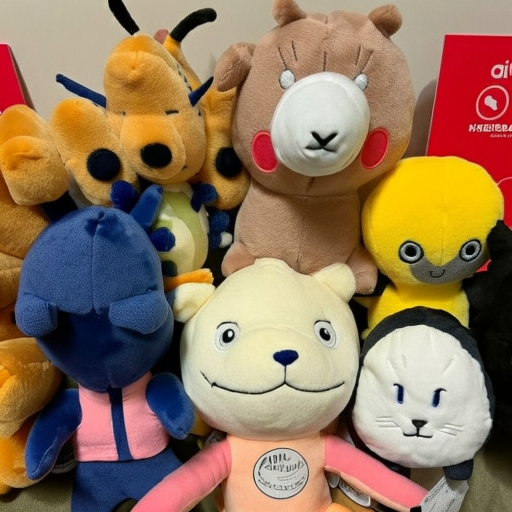}
    \includegraphics[width=0.23\textwidth]{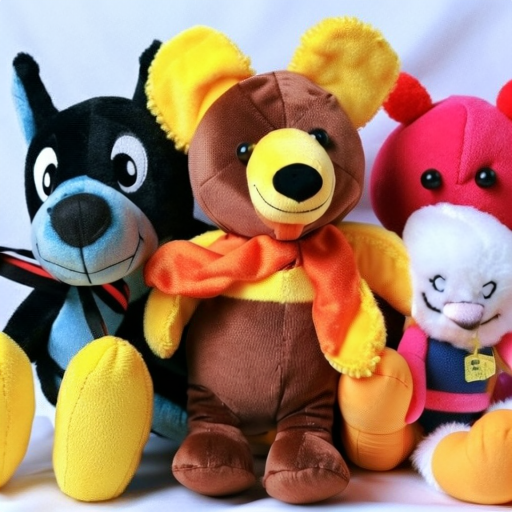}
    \caption{\textit{A group of stuffed animals are arranged together.}}
\end{subfigure}
\caption{\textbf{Variation with seeds: T2I Results on MS-COCO with MMDiT-S}. (left-to-right) We provide results of images generated from the given text prompt with different seeds (0,1,2,3) with our method.}
\label{fig:large-ms-coco-visualization-diff-seeds}
\end{figure}

In~\Cref{fig:large-ms-coco-visualization-diff-seeds} and in~\Cref{fig:small-ms-coco-visualization-additional}, we present the image samples for various methods.

\end{document}